\documentclass[inte,nonblindrev]{informs3} 

\OneAndAHalfSpacedXII 

\usepackage[caption=false]{subfig}

\TheoremsNumberedThrough     

\EquationsNumberedThrough    

\graphicspath{{Figures/}}
\usepackage{float}
\usepackage{booktabs, tabularx}

\newcolumntype{C}{>{\centering\arraybackslash}X}

\newcommand{\ignore}[1]{}
\usepackage{hyperref}
\begin{document}


\RUNAUTHOR{Blom, Pendavingh, and Spieksma}

\RUNTITLE{Filling a theatre in times of corona}

\TITLE{Filling a theatre in times of corona}

\ARTICLEAUTHORS{%
\AUTHOR{Danny Blom, Rudi Pendavingh, Frits Spieksma}
\AFF{Department of Mathematics and Computer Science, Eindhoven University of Technology, The Netherlands, \EMAIL{d.a.m.p.blom@tue.nl, r.a.pendavingh@tue.nl, f.c.r.spieksma@tue.nl,}}
} 

\ABSTRACT{%
In this paper, we introduce an optimization problem posed by the Music Building Eindhoven (MBE) to deal with the economical consequences of the COVID-19 pandemic for theatre halls. We propose a model for maximizing the number of guests in a theatre hall that respects social distancing rules, and is based on trapezoid packings. Computational results show that up to 40\% of the normal capacity can be used for a single show setting, and up to 70 \% in case artists opt for two consecutive performances per evening.
}%





\maketitle

\section{Prologue}

All around the world, the corona-crisis has hit the cultural sector hard. Festivals are cancelled, orchestra's are at the brink of bankruptcy, choirs have stopped performing, and theatres are struggling to survive. Different countries, or regions, have imposed different rules in an attempt to stop the spread of the virus. We do not aim here to overview the precise (dynamic!) contents of all these rules, and their impact on the cultural sector; a number of descriptions of such rules and their impact can be found on governmental websites (e.g.~\cite{aus}, \cite{germany}, ~\cite{sweden}, ~\cite{uk} and ~\cite{usa}), and in other contributions (such as~\cite{jacobs}). 

The situation in the Netherlands is not atypical from other countries or regions. Starting March 12, 2020 until June 1, 2020 all performances were cancelled or suspended. From June 1 onwards, a relaxation of the rules has allowed performances with at most thirty guests, as long as non-family members were seated at least 1.5  meters apart. The upper bound on the number of guests for indoor performances was eventually increased to one hundred on July 1, 2020; a description of the current rules can be found at \cite{nl}. Clearly, these rules have a dramatic impact on the operation of any theatre, and despite governmental efforts theatres are struggling to survive. As a consequence, many employees in this sector risk losing their jobs.

Indeed, for many theatres, the challenge is to find a way to welcome their guests while satisfying the distance rules, and still be commercially viable. Many creative efforts have resulted in a number of ideas that are being experimented with (see, for example, the use of a so-called nebulizer device, see~\cite{nebulizer}). Here, we focus on the question to what extent large audiences can still be accommodated in a theatre when distance rules must be satisfied. We describe a mathematical model that, given the layout of the seats in a theatre and the distribution of the demand, computes a safe seating arrangement that attains the maximum occupation of the theatre. 

The Music Building Eindhoven (MBE), located in the city of Eindhoven in the Netherlands, features a ``Grand Room'' (1250 seats) and a ``Small Room'' (400 seats). This theatre has served as a motivation for this study, and all our computational efforts are based on its two rooms. Our findings have been implemented by the MBE, allowing them to remain open.

In Section~\ref{sec:1.5} we give a precise problem description, and in Section~\ref{sec:trapezoid} we phrase the problem in terms of packing of trapezoids. In Section~\ref{sec:model} we give our model, and in Section~\ref{sec:results} we show solutions of the model on instances coming from the MBE. Upper bounds are discussed in Section~\ref{sec:ub}, and we conclude in Section~\ref{sec:conclusion}.

\section{Problem description}
\label{sec:1.5}

Here, we describe the crucial ingredients of our problem. Seats, distances and forbidden zones are discussed in Section~\ref{sec:seats+}, and target profiles are explained in Section~\ref{sec:target}. This allows us to arrive at a problem statement given in Section~\ref{sec:problemstatement}.

\subsection{Seats, distances and forbidden zones}
\label{sec:seats+}

When a theatre wants to offer a corona-proof experience to its customers, a few constraints need to be taken into account. Obviously, safety is of utmost importance and therefore, the subset of seats that could be used for reservations needs to be chosen according to the guidelines provided by the government. We realize that these guidelines vary for different countries. However, a common denominator between different countries is that members of distinct families (or {\em bubbles}) should keep a prespecified distance from each other to prevent the spread of the COVID-19 virus. In the Netherlands, two people should be seated at least 1.5 meter apart (unless they are members from the same household), as established by the Dutch Government~\cite{nl}.

\begin{figure}[H]
	\FIGURE
	{\includegraphics[height = 5cm, width= 7cm]{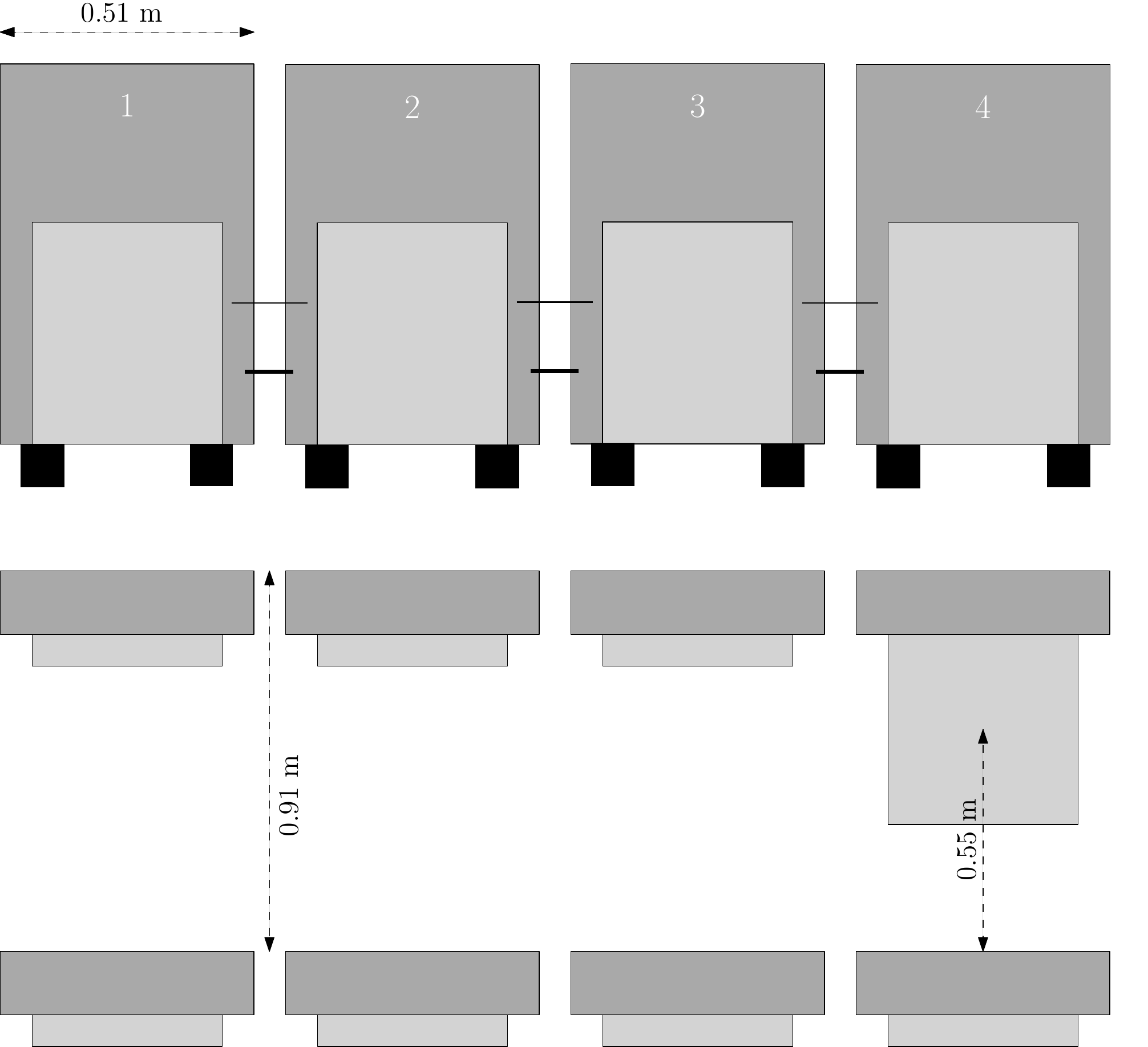}}
	{Front and upper view of four seats in MBE, with corresponding measures. The height difference of consecutive rows is 0.31m, due to acoustics and visibility reasons.\label{fig:afmetingen}}
	{}
\end{figure}

Figure~\ref{fig:afmetingen} shows a sketch of four consecutive seats, viewed from the front and from above, and the corresponding interseat distances of seats in the MBE. 

\begin{figure}[H]
	\FIGURE
	{\includegraphics[scale = 0.3]{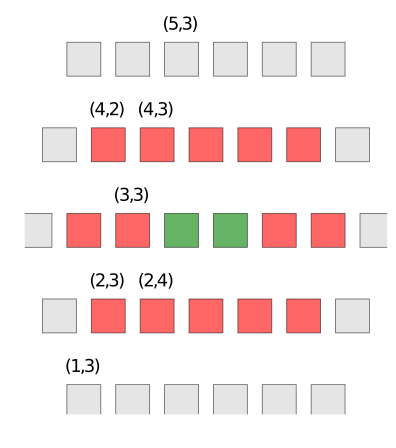}}
	{The red seats cannot be occupied whenever the green seats are occupied by a pair. The seats with seat number 3 in consecutive rows are situated on a straight line. This example is based on the measures of the MBE.\label{fig:verboden_gebied}}
	{}
\end{figure}

Due to the exception of the distance rules for family members, another relevant factor is the {\em size} $t \in T$ of a family, with $T \subset \mathbb{Z}_+$ the set of allowed family sizes. In particular, we will call a family of size 1 a {\em singleton}, of size 2 a {\em pair}, of size 3 a {\em triple}, and a family of size 4 a {\em quad}. Guests from the same household are allowed to sit next to each other, within the 1.5m bound; in fact, we assume that a family of size $t$ occupies $t$ consecutive seats. Based on the distances in Figure~\ref{fig:afmetingen}, it follows that whenever a certain seat is occupied, there is a ``ring'' of seats around it that are forbidden for use by a member of another family. The forbidden ring corresponding to a pair is depicted in red in Figure~\ref{fig:verboden_gebied}.

Consider a theatre, and let $\mathcal{S}$ denote the set of seats of the theatre. Each seat is specified by its {\em row} $r$, and its {\em position} $s$ in row $r$. Formally, a seat is a pair of integers $(r,s)\in \mathbb{Z}\times\mathbb{Z}$ and the set of seats is a collection $\mathcal{S}\subseteq \mathbb{Z}\times\mathbb{Z}$. Typically, the seats in each row are numbered starting with $s=1, 2, 3, \ldots$, so that the relative position of the seats $(r, s)$ and $(r', s)$ will depend on where rows $r$ and $r'$ start.  For the description of our model it will be convenient to assume that the seats in each row are numbered such that for each $s \in \mathbb{Z}$, the seats $\{ (r,s)\in \mathcal{S}: r\in \mathbb{Z}\}$ are in  a straight line. Figure~\ref{fig:verboden_gebied} also illustrates this convention for $\{ (r,3)\in \mathcal{S}: r\in \mathbb{Z}\}$.

Again, in a typical theatre (such as MBE), consecutive rows are shifted relative to each other (for reasons of visibility), so that the four seats $(r+1,s-1), (r+1, s),  (r-1, s), (r-1, s+1)$ form the corners of a rectangle with center $(r,s)$. This seat renumbering method is illustrated in Figure~\ref{fig:verboden_gebied} for $(r,s) = (3,3)$.
	
When we denote the distance between the centers of adjacent seats in a row by $a$, and the spacing of consecutive rows is denoted by $b$, then the distance between the centers of seats $(r, s)$ and $(r', s')$ is
$$d((r, s), (r', s'))= \sqrt{\left((s-s'+\frac{1}{2}(r-r')\right)^2a^2+ (r-r')^2b^2}.$$




Assuming that members of different families may not be seated within distance $c$, the `forbidden zone' for members of other families surrounding a person in seat $(0,0)$ is
$$\mathcal{F}:=\{(r,s): d((r,s), (0,0))<c\}.$$
Taking distances $a=0.51m$ and $b=0.95m$ as in Figure \ref{fig:afmetingen}, and a forbidden distance of $c=1.5m$, a calculation reveals that $\mathcal{F}$ is a collection of 13 seats: the occupied seat itself, two seats on each side in in the same row, and four nearby seats in each adjacent row. In general, the forbidden zone for a family of size $t$ consists of $13 + 2(t-1) = 2t+11$ seats. In Figure~\ref{fig:verboden_gebied}, we indeed see that the forbidden zone of a pair consists of 15 seats, as there is an extra forbidden seat in both adjacent rows. 

The forbidden zone surrounding each other seat $(r,s)$ is just a shifted version of this forbidden zone around $(0,0)$:
$$\mathcal{F}_{r,s}:=\{(r,s)\}+\mathcal{F}=\{(r+r', s+s'): (r',s')\in \mathcal{F}\}.$$
Members of the same family are allowed to be in each other's forbidden zone, but the union of their forbidden zones is forbidden for members of other families. A family of size $t$ located at $(r,s)$ will occupy the seats
$\mathcal{S}_{r,s,t}:=\{(r,s+i): i=0, \ldots, t-1)\}$ and will have a forbidden zone
$$\mathcal{F}_{r,s,t}:=\mathcal{S}_{r,s,t}+\mathcal{F}=\{(r+r', s+s'+i): i=0, \ldots, t-1, (r', s')\in \mathcal{F}\}.$$ 
(Notice that we use the Minkowski sum when adding two sets $A$ and $B$, i.e. $A+B = \{a+b:~a \in A, b \in B\}$.)
A subset $\mathcal{A}\subseteq \mathcal{S}\times T$ is a {\em seating arrangement} if each $(r,s,t)\in \mathcal{A}$ indicates a possible location of a family of size $t$ at $(r,s)$, i.e. if $\mathcal{S}_{r,s,t} \subseteq \mathcal{S}$ for each $(r,s,t)\in \mathcal{A}$ and $\mathcal{S}_{r,s,t} \cap \mathcal{S}_{r',s',t'} = \emptyset$ for each distinct pair $(r,s,t), (r',s',t') \in \mathcal{A}$.

\begin{definition}
A seating arrangement $\mathcal{A}$ is {\em safe} if 
\[\mathcal{S}_{r,s,t} \cap \mathcal{F}_{r',s',t'} = \emptyset\]
for each distinct pair $(r,s,t), (r',s',t') \in \mathcal{A}$.\end{definition}
Thus, a seating arrangement $\mathcal{A}$ is safe if no member of a family is in the forbidden zone of another family.

For any seating arrangement $\mathcal{A}$, let $n_t(\mathcal{A}):=|\{(r,s,t)\in \mathcal{A}: (r,s)\in \mathcal{S}\}|$ capture the number of families of size $t$ in $\mathcal{A}$, $t \in T$.

\begin{definition}
The {\em size} of a seating arrangement $\mathcal{A}$ is $\sum_{t \in T} t \cdot n_t(\mathcal{A})$.
\end{definition}
Thus, the size of a seating arrangement corresponds to the number of customers present.

\subsection{Target profiles}
\label{sec:target}
Apart from providing a safe environment for the audience while enjoying a performance, a theatre needs to consider its booking strategy. In general, multiple factors play a role when deciding upon such a strategy (see Baldin and Bille~\cite{balbil}, and the references contained therein). One option is to sell the individual seats (perhaps after segmentation into classes) chosen by customers in a first-come first-serve manner. The risk of such a strategy is that customers choose seats that do not lead to a maximum occupancy. Another option is to simply sell tickets, and only reveal very shortly before the start of the performance which particular seats are assigned to which individual customers. This allows the theatre flexibility to find a maximum occupancy, yet customers might find it unattractive not to be able to choose their specific seats. Without going into details of the various considerations, we have opted, in collaboration with the MBE for a policy that (i) allows customers to choose their seats, and (ii) uses a so-called {\em target profile} to take the size of families visiting the performance into account. Indeed, prior information on the distribution of the customers over singletons, pairs, triples and quads is valuable information and can serve as a proxy for customer behaviour. 

\begin{definition}
A {\em target profile} for a seating arrangement $\mathcal{A}$ is a vector $\vec{p} = (p_t)_{t \in T} \in [0,1]^T$ such that $\sum_{t \in T} p_t = 1$. 
\end{definition}
Each entry $p_t$ indicates a targeted proportion of the reservations corresponding to families of size $t$, i.e. we aim for a seating arrangement $\mathcal{A}$ for which
\begin{equation}\label{eq:target_profile}\frac{n_t(\mathcal{A})}{\sum_{t' \in T} n_{t'}(\mathcal{A})} \approx p_t,\qquad \forall t \in T.
\end{equation}

A target profile can be determined through statistical analysis or machine learning models applied on historical data. We show in Section~\ref{sec:problemstatement} how we formalize this aim. 


\subsection{Problem statement}
\label{sec:problemstatement}
When we use the target profile to proxy customer behaviour as input, we can describe the problem in the following way:\\
\newline
\textbf{Problem:} MAXIMUM PROFILED SEATING ARRANGEMENT\\
\textbf{Instance:} a tuple $(\mathcal{S}, T, \vec{p}, \epsilon)$ consisting of a set of seats $\mathcal{S} \subseteq \mathbb{Z} \times \mathbb{Z}$, a set of allowed family sizes $T \subset \mathbb{N}_+$, 
a target profile $\vec{p} = (p_t)_{t \in T}$ and $\epsilon \in (0,1]$.\\
\textbf{Goal:} Find a safe seating arrangement $\mathcal{A}$ of maximum size
such that the following conditions hold: 
\[(p_t - \epsilon)\sum_{t' \in T} n_{t'}(\mathcal{A}) \le n_t(\mathcal{A}) \le (p_t + \epsilon)\sum_{t' \in T} n_{t'}(\mathcal{A}), \quad \forall t \in T.\]

Notice that there is no unique way to model a condition as provided in Equation (\ref{eq:target_profile}). One alternative is to consider strict lower bounds on the number $n_t(\mathcal{A})$ of families of size $t$. Furthermore, $\epsilon$ is used as a multiplicative threshold parameter, but one could also use it as an additional parameter instead. In the following section, we describe a nontrivial connection between finding a safe seating arrangement and the problem of packing a maximum weight set of trapezoids.

\section{Trapezoid packings}
\label{sec:trapezoid}
We describe in Section~\ref{sec:seatingtrapezoid} a nontrivial connection between finding a safe seating arrangement and the problem of packing a maximum number of trapezoids in a polygonal shape, and in Section~\ref{sec:bound} we prove a bound on the number of families that fit in a theatre. In Section~\ref{sec:hilbert} we analyze, as an intermezzo, a theatre with an infinite number of rows having each an infinite number of seats, and in Section~\ref{sec:largetheatres} we discuss the occupation density of seating arrangements in large theatres.

\subsection{From safe seating arrangements to trapezoid packings}
\label{sec:seatingtrapezoid}

Given the structure of the forbidden zone $\mathcal{F}$, we are able to formulate a model based on {\em trapezoids}. The {\em trapezoid} based at $(0,0)$ is the collection of seats $$\mathcal{T}:=\{(0,-1), (0,0), (0,1), (1, -1), (1, 0)\}.$$ 
The trapezoid based at $(r,s)$ then is  $$\mathcal{T}_{r,s}:=\{(r,s)\}+\mathcal{T}=\{(r+u, s+v): (u,v)\in \mathcal{T}\},$$
and the trapezoid of a family of size $t$ located at $(r,s)$ is
$$\mathcal{T}_{r,s,t}:=\mathcal{S}_{r,s,t}+\mathcal{T}=\{(r+u, s+v+i): i=0, \ldots, t-1, (u, v)\in \mathcal{T}\}.$$
\begin{figure}[H]
	\FIGURE
	{\includegraphics[scale = 0.3]{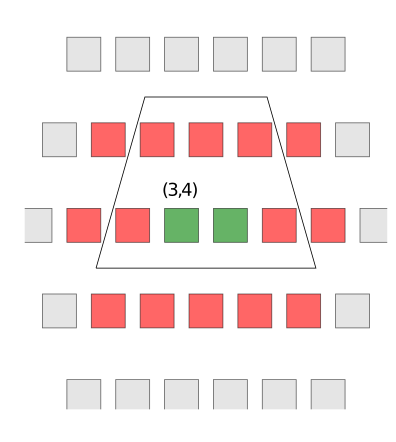}}
	{The trapezoid $\mathcal{T}_{3,4,2}$ together with its associated forbidden zone $\mathcal{F}_{3,4,2}$ (given by red and green seats).\label{fig:trapezoid}}
	{}
\end{figure}
The trapezoid $\mathcal{T}$ is chosen so that
\begin{equation}\label{minkowski}
\mathcal{F}= \mathcal{T}+(-\mathcal{T}),
\end{equation}
where $\mathcal{T}+(-\mathcal{T}):=\{(u,v)-(u',v'): (u,v), (u', v')\in \mathcal{T}\}$. 
This key property will allow us to show:
\begin{theorem} \label{thm:trap}Let $\mathcal{A}\subseteq \mathcal{S}\times T$ be a seating arrangement. Then $\mathcal{A}$ is safe if and only if 
$$\{\mathcal{T}_{r,s,t}: (r,s,t)\in\mathcal{A}\}$$
is a collection of pairwise disjoint trapezoids.\end{theorem}

Theorem~\ref{thm:trap} forms the basis of the integer programming models in Section~\ref{sec:model}, which solve the main problem of this paper for the MBE, and which in principle apply to theatres $\mathcal{S}$ of any given shape or irregular form. Theorem \ref{thm:trap} also enables us to analyse, in the remainder of this section, the limiting behaviour of optimal arrangements for large 'square' theatres.

The proof of Theorem~\ref{thm:trap} takes the form of 2 lemma's.
\begin{lemma} Let $(r,s), (r',s')\in \mathcal{S}$. Then 
$(r,s)\in \mathcal{F}_{r',s'} ~\Longleftrightarrow~ \mathcal{T}_{r,s}\cap \mathcal{T}_{r',s'}\neq\emptyset.$\end{lemma}
\proof{Proof.} Necessity: Suppose $(r,s)\in \mathcal{F}_{r',s'}$. Since 
$\mathcal{F}_{r',s'}=(r',s')+\mathcal{F}=(r',s')+\mathcal{T}+(-\mathcal{T})$ by \eqref{minkowski}, it follows that there are $(u, v), (u', v')\in \mathcal{T}$ so that $(r,s)=(r', s')+(u', v')-(u,v)$. 
Then $$\mathcal{T}_{r,s}\ni (r,s)+(u,v)=(r',s')+(u',v')\in \mathcal{T}_{r',s'},$$
so that 
$\mathcal{T}_{r,s}\cap \mathcal{T}_{r',s'}\neq\emptyset$, as required.

Sufficiency: Suppose $\mathcal{T}_{r,s}\cap \mathcal{T}_{r',s'}\neq\emptyset$. Then $(r,s)+(u,v)=(r',s')+(u',v')$ for some $(u,v), (u',v')\in \mathcal{T}$. Then 
$(r,s)=(r',s')+(u',v')-(u,v)\in (r,s)+\mathcal{T}+(-\mathcal{T})=\mathcal{F}_{r',s'}$, as required.\Halmos\endproof

\begin{lemma} Let $(r,s,t), (r',s',t')\in \mathcal{S}\times T$. Then 
$$\mathcal{S}_{r,s,t}\cap \mathcal{F}_{r',s',t'}\neq \emptyset ~\Longleftrightarrow~ \mathcal{T}_{r,s,t}\cap \mathcal{T}_{r',s',t'}\neq\emptyset.$$\end{lemma}
\proof{Proof.} We have $\mathcal{S}_{r,s,t} \cap  \mathcal{F}_{r',s',t'}\neq \emptyset$ if and only if there are $i\in \{0,\ldots, t-1\}, i'\in \{0,\ldots, t'-1\}$ so that $(r, s+i)\in \mathcal{F}_{r', s'+i'}$. By the previous lemma, this is equivalent to 
$$\mathcal{T}_{r,s+i}\cap \mathcal{T}_{r',s'+i'}\neq\emptyset\text{
for some }i\in \{0,\ldots, t-1\}, i'\in \{0,\ldots, t'-1\}.$$ 
In turn, this is equivalent to $\mathcal{T}_{r,s,t}\cap \mathcal{T}_{r',s',t'}\neq\emptyset$.\Halmos\endproof

By this lemma, we may replace the asymmetrical condition $\mathcal{S}_{r,s,t} \cap  \mathcal{F}_{r',s',t'}= \emptyset$ in the definition of a safe arrangement by the equivalent symmetrical condition $\mathcal{T}_{r,s,t} \cap  \mathcal{T}_{r',s',t'}=\emptyset$. This proves Theorem \ref{thm:trap}.

\subsection{A volume bound}
\label{sec:bound}
Under normal circumstances, it is clear how guests of a theatre use the available resources: each guest needs one seat. With social distancing rules, it is not immediately clear to what extent guests claim the resources. There will be many empty seats in any safe seating arrangement, and it is not obvious which guest to blame or charge.
Theorem \ref{thm:trap} is helpful in this sense, as it makes clear that each family with $t$ members at $(r,s)$ blocks the seats $\mathcal{T}_{r,s,t}\cap \mathcal{S}$, and so is responsible at least for the emptiness of these seats.
If $(r,s)$ is sufficiently far away from the boundary of the theatre, so that $\mathcal{T}_{r,s,t}\subseteq \mathcal{S}$, this amounts to blocking $|\mathcal{T}_{r,s,t}|=2t+3$ seats. Ignoring the boundary effect, this gives a rough upper bound on the number of families of size $t$ that can fit the theatre safely: $|\mathcal{S}|/(2t+3)$.

The following consequence of Theorem \ref{thm:trap} describes how we may take the boundary of a collection of seats $\mathcal{S}$ into account when estimating the capacity of a theatre.
\begin{theorem}\label{thm:volume_bound} Let $\mathcal{A}$ be a safe seating arrangement in $\mathcal{S}$. We have:
$$\sum_{t=1}^\infty (2t+3)n_t(\mathcal{A})\leq |\mathcal{S}+\mathcal{T}|.$$
\end{theorem}
\proof{Proof.} For each $(r,s,t)\in \mathcal{A}$, the family of size $t$ which is located at $(r,s)$ will occupy the seats
$\mathcal{S}_{r,s,t}\subseteq \mathcal{S}$ and hence for the corresponding trapezoid we have
$$\mathcal{T}_{r,s,t}=\mathcal{S}_{r,s,t}+\mathcal{T}\subseteq \mathcal{S}+\mathcal{T}.$$
By Theorem \ref{thm:trap}, each seat of $\mathcal{S}+\mathcal{T}$ is in at most one of trapezoids $\{\mathcal{T}_{r,s,t}: (r,s,t)\in\mathcal{A}\}$, and hence
$$ |\mathcal{S}+\mathcal{T}|\geq \sum_{(r,s,t)\in \mathcal{A}} |\mathcal{T}_{r,s,t}|=\sum_{t} (2t+3)n_t(\mathcal{A}),$$
as required.\Halmos\endproof

Note that the set $(\mathcal{S}+\mathcal{T})\setminus \mathcal{S}$ consists of a rim of seats adjacent to the
boundary of $\mathcal{S}$ in $\mathbb{Z}\times\mathbb{Z}$. 

For large theatres with a relatively simple boundary, the collection $\mathcal{S}+\mathcal{T}$ of the theorem is only marginally larger than $\mathcal{S}$. 
For example, let  $\mathcal{S}^k$ be a block of $k$ rows of $k$ seats each, then $|\mathcal{S}^k|=k^2$ and $|\mathcal{S}^k+\mathcal{T}|\leq (k+1)(k+2)$. Then $$\frac{|\mathcal{S}^k+\mathcal{T}|}{|\mathcal{S}^k|}\rightarrow 1$$
as $k\rightarrow \infty$. 

We analyse the limiting case of this sequence in Section~\ref{sec:hilbert}, and then we return to safe arrangements in large theatres $\mathcal{S}^k$ in Section~\ref{sec:largetheatres}. Though the square theatre $\mathcal{S}^k$ is perhaps artificial, very large venues such as sport stadiums are sufficiently similar to merit comparison. 

\begin{figure}[H]
	\FIGURE
	{\includegraphics[scale = 0.08]{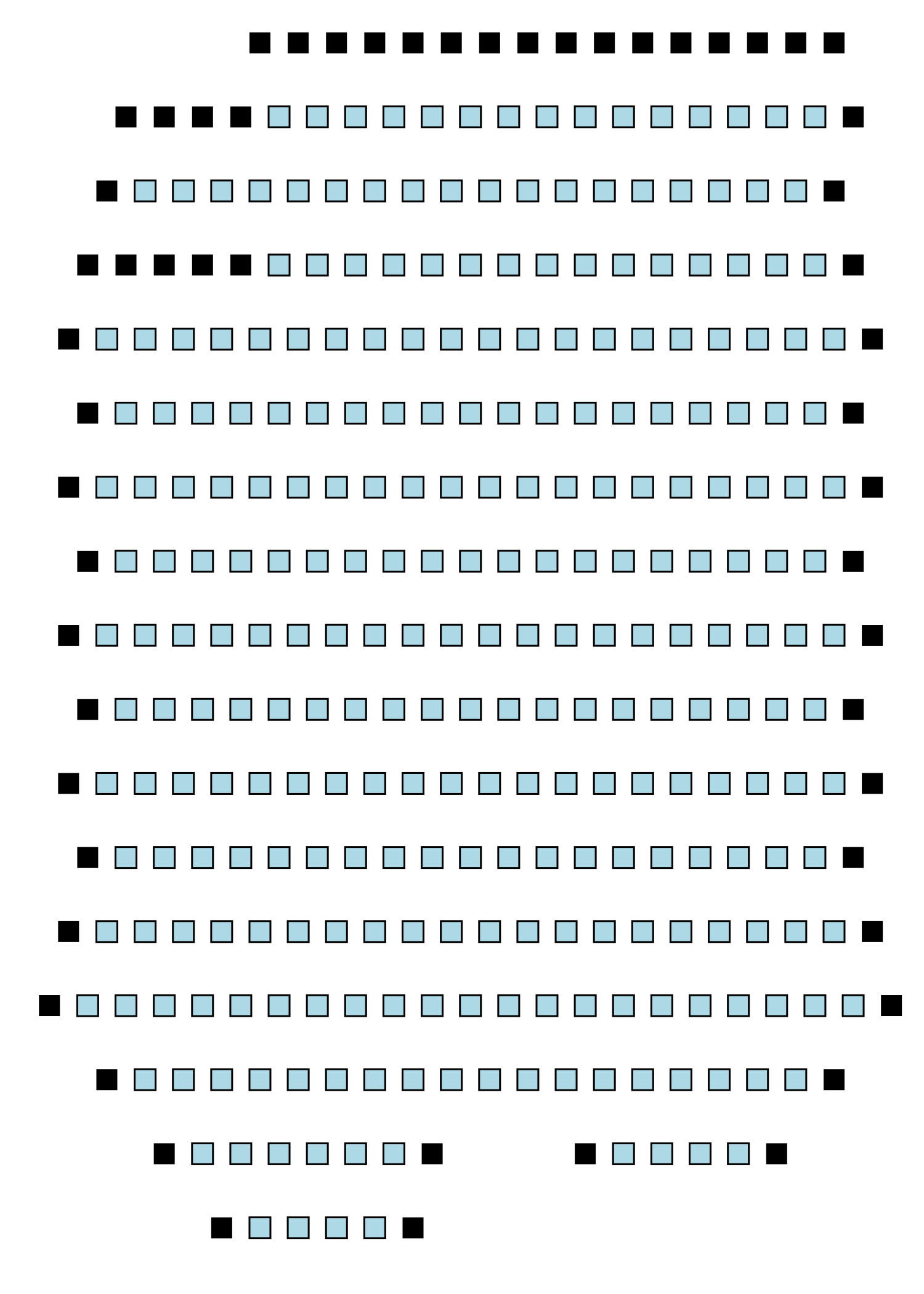}}
	{Floor plan of the ground floor of the Grand Room of the MBE. The set of seats $\mathcal{S}$ is colored lightblue, with the virtual rim of seats $(\mathcal{S}+\mathcal{T})\setminus \mathcal{S}$ colored in black. \label{fig:virtual_seats}}
	{}
\end{figure}

\subsection{Intermezzo: the Hilbert theatre}
\label{sec:hilbert}
Inspired by the famous thought experiment of Hilbert on the concept of infinity, the {\em Hilbert theatre} has seats $\mathcal{S}^\infty=\mathbb{Z}\times\mathbb{Z}$: it is an infinite sea of regularly spaced seats. 

For each family size $t$, the seating arrangement
$$\mathcal{A}^t:=\{u(2,-1)+v(1, t+1): u,v\in \mathbb{Z}\}$$
is such that the corresponding collection of trapezoids
$\{\mathcal{T}_{r,s,t}: (r,s,t)\in\mathcal{A}^t\}$
covers each seat in $\mathcal{S}^\infty$ exactly once. 

\begin{figure}[H]
	\FIGURE
	{\includegraphics[scale = 0.15]{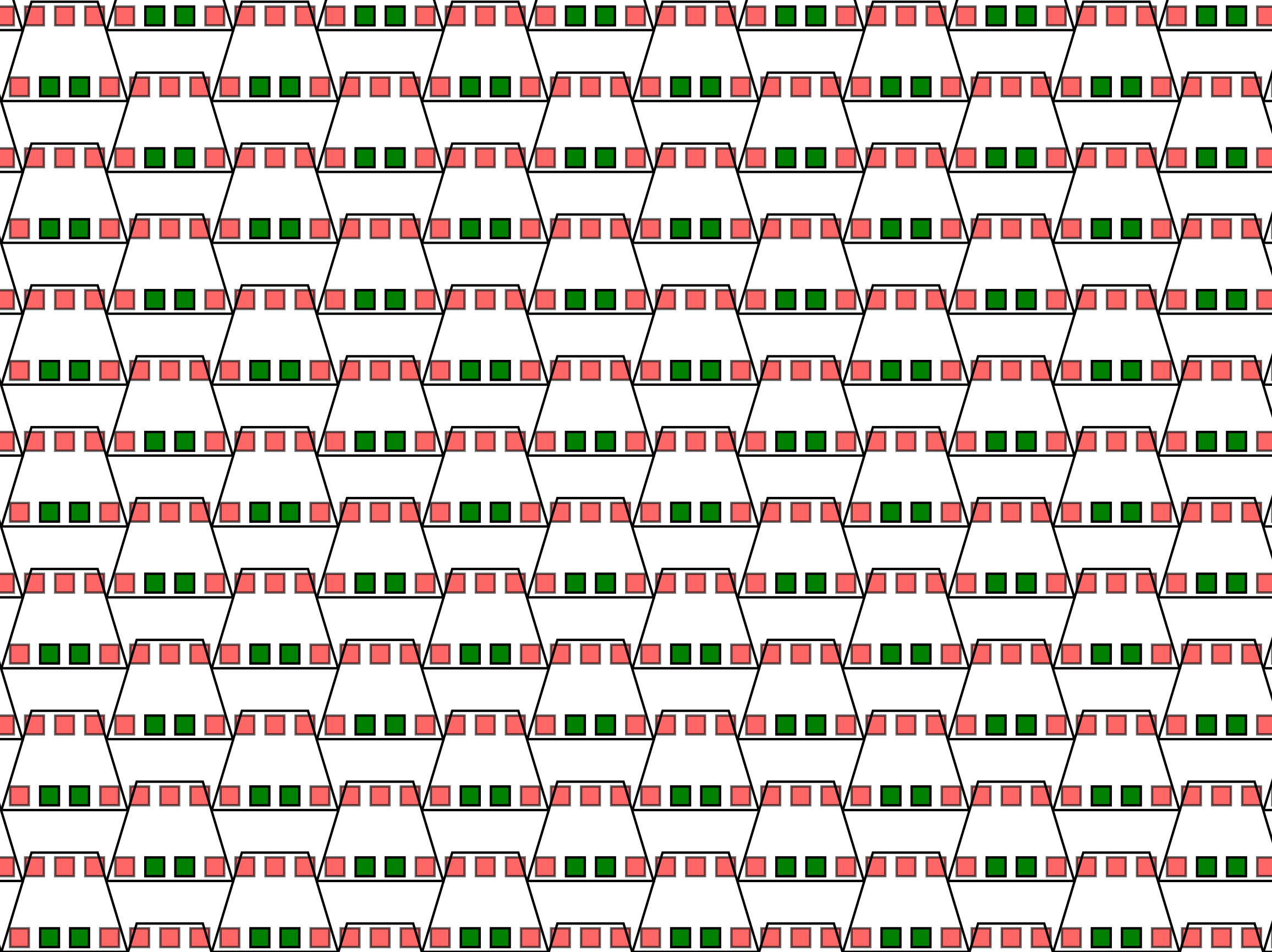}}
    {The safe seating arrangement $\mathcal{A}^2$ in the Hilbert theatre.\label{fig:hilbert_theatre}}
    {}
\end{figure}

 Hence $\mathcal{A}^t$ is a safe seating arrangement, and the average density of occupied seats in $\mathcal{A}^t$ equals the proportion of occupied seats within each trapezoid
$$\frac{|\mathcal{S}_{r,s,t}|}{|\mathcal{T}_{r,s,t}|}= \frac{t}{2t+3}=\frac{1}{2}-\frac{3}{4t+6}=:d_t.$$
We cannot hope to attain a better density than $d_t$ in any arrangement with families of size $t$. This shows that in the Hilbert theatre, the maximum density when packing families of size $t$ is $d_t$; notice that this value increases with $t$ and will never exceed $\frac{1}{2}$.
The following table shows the density $d_t$ and its reciprocal $1/d_t$  for small family sizes $t$.
$$\begin{array}{l | ccccccc| r}
t & 1& 2& 3& 4& 5& 6& \cdots & \infty\\
\hline
d_t& 0.20 & 0.29 & 0.33 & 0.36 & 0.38 & 0.4 &\cdots & 0.5\\
1/d_t& 5 & 3.5 & 3 & 2.75 & 2.6 & 2.5 & \cdots&2
\end{array}$$
As the table shows, the minimum use of seats per family member ($1/d_t$) decreases steeply in this initial range.

\subsection{The occupation density of large theatres}\label{sec:largetheatres}
Finite theatres tend to approximate the Hilbert theatre as they become larger.
For a seating arrangement $\mathcal{A}$ in a finite theatre $\mathcal{S}$, we formally define the {\em occupation density } as $$d(\mathcal{A}):=\frac{\sum\limits_{(r,s,t)\in\mathcal{A}} t}{|\mathcal{S}|}.$$ 
 In our square theatre $\mathcal{S}^k$, there is a seating arrangement of families of size $t$ which arises by restricting $\mathcal{A}^t$ to the locations available in $\mathcal{S}^k$:
$$\mathcal{A}^{t,k}:=\{(r,s,t)\in \mathcal{A}^t: \mathcal{S}_{r,s,t}\subseteq \mathcal{S}^k\}.$$
Then each row of $\mathcal{S}^k$ will see at least $\lfloor\frac{k}{2t+3}\rfloor$ families in $\mathcal{A}^{t,k}$, and hence $\mathcal{A}^{t,k}$ attains an overall density of occupied seats of at least
$$d\left(\mathcal{A}^{t,k}\right)=tk\lfloor\frac{k}{2t+3}\rfloor/k^2\geq t\left(\frac{k}{2t+3}-1\right)/k=d_t-\frac{t}{k}.$$
Evidently, this lower bound on the density tends to $d_t$ as $k\rightarrow \infty$.

For an upper bound on the occupation density of any arrangement $\mathcal{A}$ in $\mathcal{S}^k$ consisting families of size $t$ only, we may bound the number $n_t$ of families in $\mathcal{A}$ by applying Theorem~\ref{thm:trap}. This gives
$$(2t+3)n_t\leq |\mathcal{S}^k+\mathcal{T}|\leq (k+1)(k+2).$$
Since each family has $t$ members and the total number of seats in $\mathcal{S}^k$ is $k^2$,  we obtain an upper bound on the occupation density of
$$d(\mathcal{A})=\frac{t n_t}{k^2}\leq \frac{t(k+1)(k+2)}{(2t+3)k^2}\leq d_t\left(1+\frac{3k+3}{k^2}\right).$$
So the occupation density of such $\mathcal{A}$ may marginally exceed the density $d_t$ of $\mathcal{A}^t$ for low values of $k$, but the upper bound tends to $d_t$ as $k\rightarrow \infty$. In particular, we find that the density of $\mathcal{A}^{t,k}$ tends to $d_t$ as $k\rightarrow \infty$, since both lower and the upper bound converge to this value. 
So, the upper bound from Theorem~\ref{thm:volume_bound} ultimately dictates the maximum density of arrangements in sufficiently large theatres. This extends to arrangements where the relative proportion of families of size $t$ is restricted, in the following precise sense. For an arrangement $\mathcal{A}$ of $\mathcal{S}$, let $p(\mathcal{A}):\mathbb{N}\rightarrow \mathbb{R}$ record the relative proportion of families of size $t$:
$$ p(\mathcal{A}): t \mapsto n_t(\mathcal{A})/|\mathcal{A}|.$$
For any $p:\mathbb{N}\rightarrow \mathbb{R}_+$, put $D(p):=\sum_t p_t d_t$.
\begin{theorem}\label{thm:bound2} For any safe arrangement $\mathcal{A}$ of $\mathcal{S}$, we have 
$$d(\mathcal{A})\leq D(p(\mathcal{A}))\frac{|\mathcal{S}+\mathcal{T}|}{|\mathcal{S}|}.$$
Moreover, for any $p:\mathbb{N}\rightarrow \mathbb{R}_+$ such that $\sum_t p_t =1$, there exists a sequence of arrangements $\mathcal{A}^k$  of $\mathcal{S}^k$ such that 
$p(\mathcal{A}^k)\rightarrow p$ and $d(\mathcal{A}^k)\rightarrow D(p)$ as $k\rightarrow \infty$.
\end{theorem}
We omit the proof, noting that it is a straightforward extension of the argumentation above.

\paragraph*{Remark on special seating arrangements}
The highest possible occupation densities are attained by arrangements where there are families placed in each row of the theatre (see Section~\ref{sec:results}). 
For comparison, it is interesting to consider safe seating arrangements in $\mathcal{S}^k$ where each occupied row is sandwiched between two empty rows. If $\mathcal{A}$ is such an arrangement with families of size $t$, then $\mathcal{A}$ is safe if and only if there are two empty seats between any two 
families in the same row. Restricting to such simpler arrangements may have practical advantages. It will be easier to guide guests to their seats, and in fact it becomes possible to rearrange the order in which families take place in their row on the fly. This may translate to a lesser need for personnel hosting guests. Also, since safety of an arrangement depends only on a condition within each row, finding safe arrangements becomes so easy that it can be done manually or with software applying straightforward strategies. Without the need to incorporate a 'black box' advanced solver in the process of selling seats, the flexibility of this process may be greatly increased.

Let us analyze the densities $d'_t$ that can be obtained for such special seating arrangements for a fixed $t$. Since $\mathcal{A}$ will have $\lceil k/2\rceil$ occupied rows, we have 
$$d'_t(1-o(1))=\frac{1}{k^2}\lceil\frac{k}{2}\rceil t\lfloor\frac{k+2}{t+2}\rfloor
\leq d(\mathcal{A})\leq \frac{1}{k^2}\lceil\frac{k}{2}\rceil t\lceil\frac{k+2}{t+2}\rceil= d'_t(1+o(1))$$
as $k\rightarrow \infty$, where $d'_t:= t/(2t+4)$. This is worse than $d_t= t/(2t+3)$, but not by much. 


To assess the loss of density when using these special seating arrangements, we include a table with the densities $d'_t$, the inverse densities $1/d'_t$, and the relative densities $d'_t/d_t$.
$$\begin{array}{l | ccccccc|r}
t & 1& 2& 3& 4& 5& 6& \cdots & \infty\\
\hline
d'_t& 0.17 & 0.25 & 0.30 & 0.33 & 0.36 & 0.38 &\cdots & 0.5\\
1/d'_t& 6 & 4 & 3.33 & 3 & 2.8 & 2.67 & \cdots&2\\
d'_t/d_t& 83\% & 88\% & 90\% & 92\% & 93\% & 94\% & \cdots&100\%
\end{array}$$
\ignore{
Thus our modelling of the problem of finding safe seating arrangements in terms of trapezoids allows us to make a first rough analysis that applies to large and regular theatres. In conclusion:
\begin{itemize}
\item higher occupation densities are attained by packing larger families, with families of size $t$ attaining a density of $\approx d_t=\frac{1}{2}-\frac{3}{4t+6}$ occupied seats per available seat; 
\item as theatres are larger, the optimal packings tend to approximate this density; and
\item the packings that arise by leaving a row empty between each occupied row are suboptimal, but not by much.
\end{itemize}
}

\paragraph*{Remark on a shorter safe distance} The above analysis can be straightforwardly adapted to a safe distance of 1 meter. Then the forbidden zone becomes 
$$\mathcal{F}:=\{(-1, 0), (-1, 1), (0,-1), (0,0), (0,1), (1,-1), (1,0)\}.$$
Taking $\mathcal{T}:=\{ (0,0), (0,1), (1,0)\}$
we will again have $\mathcal{F}=\mathcal{T}+(-\mathcal{T})$, i.e. \eqref{minkowski}. Then Theorem \ref{thm:trap} holds, and we obtain
$|\mathcal{T}_{r,s,t}|=2t+1$ for each $t\in \mathbb{N}$ and 
$$d_t=\frac{|\mathcal{S}_{r,s,t}|}{|\mathcal{T}_{r,s,t}|}=\frac{t}{2t+1}.$$
There are safe seating arrangements in the Hilbert theatre that attain this density, and Theorem \ref{thm:bound2} remains valid. For suboptimal seating arrangements with empty rows between occupied rows, we obtain the densities $d'_t=\frac{t}{2t+2}$.

We have so far used our model to investigate a square and sufficiently large theatre $\mathcal{S}^k$. 
We will next explain how our model yields a computational strategy to find optimal packings for any specific theatre.

\section{An integer programming model to maximize the size of a seating arrangement}
\label{sec:model}
We describe our model in Section~\ref{sec:building}, and in Section~\ref{sec:shows} we show how the concept of  multiple shows can be embedded in the model. In Section~\ref{sec:speedup} we indicate how we speed up the solution process of the model.

\subsection{Building the model}
\label{sec:building}

With any collection $\mathcal{A}\subseteq\mathcal{S}\times T$, where $T\subseteq \mathbb{N}_+$ is a finite collection of allowed family sizes, we can associate a characteristic vector $y\in \{0,1\}^{\mathcal{S}\times T}$ with $y_{r,s,t}=1$ if and only if $(r,s,t)\in \mathcal{A}$.
Then $\mathcal{A}$ is a seating arrangement in $\mathcal{S}$ if and only if
\begin{equation}\label{cond:S}
 y_{r,s,t}=0 \qquad\text{whenever }\mathcal{S}_{r,s,t}\not\subseteq \mathcal{S}.
\end{equation}
This seating arrangement $\mathcal{A}$ is safe if and only if
\begin{equation}\label{cond:safe}
\sum_{(r',s',t'): \mathcal{T}_{r',s',t'} \ni (r,s)} y_{r',s',t'}\leq 1,\qquad \text{ for all }(r,s)\in \mathcal{S}+\mathcal{T}.
\end{equation}
From a geometric point of view, constraints (\ref{cond:safe}) ensure that each seat $(r, s) \in \mathcal{S}+\mathcal{T}$ is covered by at most one of the trapezoids it is contained in.
Finally, $\mathcal{A}$ accommodates $n_t$ families of each size $t\in T$ if
\begin{equation}\label{cond:count}
\sum_{(r,s):\mathcal{S}_{r,s,t} \subseteq \mathcal{S}} y_{r,s,t} = n_t,\qquad\text{ for each } t\in T.
\end{equation}
Thus, the feasibility of a safe seating arrangement that simultaneously accommodates $n_t$ families of size $t$ for $t\in T$ translates to an integer linear feasibility problem in variables $y_{r,s,t}$ and $n_t$.
However, without a priori conditions on the number of families of each size $t$, the optimal solutions of this problem will tend towards including many large families and few small families. This is intuitively clear, since a family of $t$ together 'wastes' a trapezoid of $2t+3$ seats, so that $2+3/t (=1/d_t)$ seats are taken per person in a family of size $t$. In the extreme case that $T$ includes large enough sizes to fill entire rows of seats with a single family, then a solution where the even rows are empty and each odd row is filled with a single family is feasible, and similar for leaving the odd rows empty and filling the even rows. One of these solutions then is optimal, and uses at least half of the seats in $\mathcal{S}$. Indeed, now that we are letting our imagination roam free, we can fill the entire theatre with a single large enough family if we also let go of our restriction that families must be seated in the same row. To ensure that we find safe seating arrangements that approximately correspond to the typical sizes of families that book seats for a performance, we use the target profile, as introduced in Section~\ref{sec:target}. Recall from the problem statement that the target profile imposes the condition
\begin{equation}\label{cond:profile}
(p_t-\epsilon)\sum_{t\in T} n_t \leq n_t\leq (p_t+\epsilon)\sum_{t\in T} n_t,\qquad\text{ for each } t\in T.
\end{equation}

In this way we obtain an integer linear program that maximizes the size of a seating arrangement over all safe seating arrangements in $\mathcal{S}$:

\begin{equation}\label{model1}\max\left\{\sum_{t\in T} tn_t: ~\eqref{cond:S}, ~\eqref{cond:safe}, ~\eqref{cond:count}, ~\eqref{cond:profile},~y\in \{0,1\}^{\mathcal{S}\times T}, ~n\in \mathbb{Z}^T\right\}.\end{equation}

Notice that the LP relaxation of \eqref{model1} gives an upper bound which, by the safety constraints \eqref{cond:safe}, is informed that each family of size $t$ occupies at least $2t+3$ seats from $\mathcal{S}+\mathcal{T}$. Evaluated with $\epsilon=0$, the LP relaxation will be at least as good as the bound of Theorem \ref{thm:bound2}.

\subsection{Consecutive shows}
\label{sec:shows}


One of the ideas that the MBE has implemented to remain commercially viable is to perform the same show during the same evening twice, each time for a different audience. We refer to this phenomenon as {\em consecutive shows}. Clearly, this puts a burden on the performing artist(s); in many cases however, this is a realistic option. The MBE, however, is not able to clean the seats in between the shows. This creates an interdependence between the two seating arrangements for each individual show as each seat can be used at most once in each of the two seating arrangements.

However, it is relatively straightforward to extend our model to find $k$ consecutive seating arrangements $\mathcal{A}_v$ for $v\in V = \{1,\ldots, k\}, k \in \mathbb{Z}_+$, so that no seat is used in two different arrangements, i.e. if $v, v'\in V$ are distinct, then
$$\mathcal{S}_{r,s,t}\cap \mathcal{S}_{r',s',t'}=\emptyset$$
for all $(r,s,t)\in \mathcal{A}_v$ and $(r',s',t')\in \mathcal{A}_{v'}$.

To model the problem of finding such consecutive seating arrangements, we use binary variables $y\in \{0,1\}^{\mathcal{S}\times T\times V}$, and integer variables $n\in \mathbb{Z}^T$.
The condition that each $\mathcal{A}_v$ is a seating arrangement of $\mathcal{S}$ becomes
\begin{equation}\label{cond:S2}
 y_{r,s,t,v}=0, \qquad\text{whenever }\mathcal{S}_{r,s,t}\not\subseteq \mathcal{S}.
\end{equation}
Safety of each  $\mathcal{A}_v$ is modelled by
\begin{equation}\label{cond:safe2}
\sum_{(r',s',t'):\mathcal{T}_{r',s',t'} \ni (r,s)} y_{r',s',t',v}\leq 1,\qquad \text{ for all }(r, s)\in \mathcal{S}+\mathcal{T}, v\in V.
\end{equation}
We also need to ensure that no seat is used more than once.
\begin{equation}\label{cond:seat2}
\sum_{v \in V}\sum_{(r',s',t'):\mathcal{S}_{r',s',t'} \ni (r,s)} y_{r',s',t',v}\leq 1,\qquad \text{ for all }(r, s)\in \mathcal{S}.
\end{equation}
Letting the $n_t$ count the overall number of families of size $t$ is accomplished by writing
\begin{equation}\label{cond:count2}
\sum_{v\in V} \sum_{(r,s)\in \mathcal{S}} y_{r,s,t,v} = n_t,\qquad\text{ for each } t\in T.
\end{equation}
The profiling condition \eqref{cond:profile} need not change at all.

Maximizing the number of guests in consecutive arrangements in $\mathcal{S}$,  whilst respecting a profile $p\in \mathbb{R}^T$ up to a fixed $\epsilon>0$, is then modelled as the following ILP:
\begin{equation}\label{model2}\max\left\{\sum_{t\in T} tn_t: ~\eqref{cond:profile}, ~\eqref{cond:S2}, ~\eqref{cond:safe2}, ~\eqref{cond:seat2}, ~\eqref{cond:count2},~ y\in \{0,1\}^{\mathcal{S}\times T\times V}, ~n\in \mathbb{Z}^T\right\}.\end{equation}

This model is rather flexible: many additional wishes can be formulated. For instance, upper bounds on $n_t$ for some $t \in T$, or a balance between the distribution in different shows, or specific (monetary) weights to maximize the revenue that could be gained, seats can all be arranged through standard modifications of the integer linear program.

\subsection{Speeding up the solution process}
\label{sec:speedup}


For some instances of ILP formulation (\ref{model2}), the corresponding LP relaxation leads to long running times of the solver. Hereunder we propose two methods that ameliorate solver performance, namely (i) adding a class of valid inequalities to strengthen the LP relaxation and (ii) using a symmetry breaking method for formulation (\ref{model2}).

\subsection*{Strengthening the linear relaxation}
The ILP formulations (\ref{model1}) and (\ref{model2}) only take into account for each individual seat $(r,s) \in \mathcal{S}$ the trapezoids $\mathcal{T}_{r',s',t'}$ that contain $(r,s)$. Using the adjacency of seats, Lemma~\ref{lemma:valid_ineq} finds a set of new valid inequalities.
\begin{lemma}\label{lemma:valid_ineq}
Let $(r,s) \in \mathcal{S}$ such that $X = \{(r,s), (r+1,s-1), (r+1,s)\} \subseteq \mathcal{S}$. Then, for each $t \in T$:
\begin{equation}
\label{ineqvalid}\sum_{(r,s,t): |\mathcal{T}_{r,s,t} \cap X| \ge 2} y_{r,s,t,v} \le 1,\qquad v\in V.\end{equation}
\end{lemma}
Notice that $X$ consists of three 
seats. 
Thus, from all trapezoids that contain at least two seats from $X$, one can select at most one.

\begin{figure}[H]
	\FIGURE
	{\includegraphics[scale = 0.15]{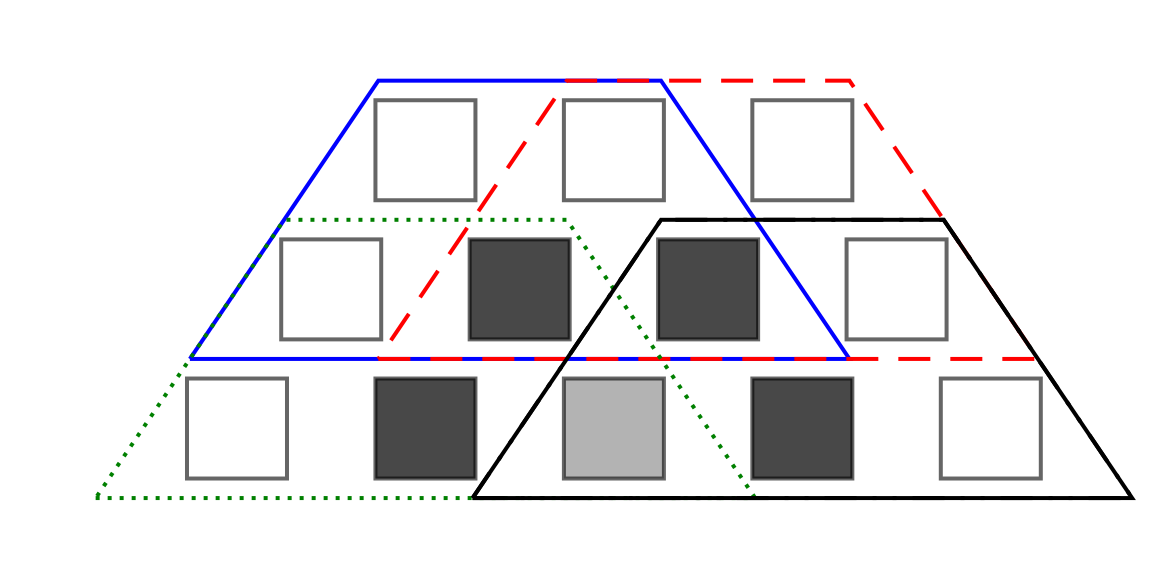}}
    {A valid inequality for this example is $y_{1,1,1} + y_{1,2,1} + y_{2,1,1} + y_{2,2,1} + y_{2,3,1} \le 1$. It separates the previously feasible LP solution $(\frac{1}{3}, \frac{1}{3}, \frac{1}{3}, 0, \frac{1}{3})$.\label{fig:strengthen_formulation}}
    {}
\end{figure}

To show that inequalities (\ref{ineqvalid}) are valid for nontrivial theatres, consider the example in Figure~\ref{fig:strengthen_formulation}, where $T = \{1\}$, $p_1 = 1$, and $V = \{1\}$. The theatre consists of a set $\mathcal{S}$ of five seats, represented by the non-white squares, with the rim of virtual white seats around it. Each of these four dark grey seats is contained in at most three trapezoids. so the LP solution with each of the four trapezoids chosen with 1/3 is feasible for (\ref{model1}), with value 4/3. However, a safe seating arrangement $\mathcal{A}$ can clearly contain at most one seat.  

\subsection*{Symmetry breaking techniques}


The presence of symmetry in a (mixed) integer programming formulation often poses a computational challenge, see e.g. Margot~\cite{margot} and Hojny and Pfetsch~\cite{hojpfe}. Indeed, naive implementations can be unsuccessful, as many equivalent problems need to be solved in the branch-and-bound procedure to ensure optimality.

Consider now a sequence of consecutive safe seating arrangements $(\mathcal{A}_v)_{v \in V}$. Then, for an arbitrary permutation $\sigma \in Sym(|V|)$, the sequence $(\mathcal{A}_{\sigma(v)})_{v \in V}$ is again feasible. The choice of permutations can even be done independently for each segment of the theatre, i.e. the ground floor and its separate balconies. The following lemma provides a class of inequalities that drastically reduces the feasible region by removing symmetries caused by these permutation groups.
\begin{proposition}\label{prop:symmetry}

Let a segment ${\ell} = 1,\ldots,n$ in a theatre have a set of seats $\mathcal{S}^{\ell} \subset \mathcal{S}$. Let $\prec$ be the standard lexicographic ordering relation on $\mathcal{S}$. For each $(r',s') \in \mathcal{S}^{\ell}, v' \in V$, a class of symmetry breaking inequalities is
\begin{equation}
\label{ineqbreaksym}
\sum_{t' \in T} y_{r',s',t',v'} \le \sum_{\substack{(r,s)\in \mathcal{S}^{\ell}\\(r,s) \prec (r',s')}}\sum_{t' \in T} y_{r,s,t',v},\qquad \forall v \in V : v < v'.
\end{equation}
\end{proposition}
Suppose we are given a seat $(r',s') \in \mathcal{S}^{\ell}, v' \in V$ and $v < v'$. The left hand side considers families of size $t$ starting on seat $(r',s')$ for show $v'$. The inequality tells us that we can only place at family starting at $(r',s')$ for show $v'$ whenever in an earlier show $v < v'$ a family was placed starting on some seat $(r,s) \prec (r',s')$. Most importantly, we claim without proof that it is a correct symmetry breaking method.
\begin{lemma}
There exists an optimal solution to (\ref{model2}) that satisfies inequalities (\ref{ineqbreaksym}).
\end{lemma}

\section{Computational results}\label{sec:results}
We implemented the models (\ref{model1}) and (\ref{model2}) above in Julia 1.3.0, using the modelling language JuMP to build the optimization model, with Gurobi as the lower level LP and MIP solver. Experiments were run on a computer equipped with an Intel Core i7-7700HQ CPU @ 2.8 GHz with 32 GB of RAM. For our experiments we considered four different target profiles $\vec{p} = \left(p_t\right)_{t \in T}$, with $T = \{1,2,3,4\}$, based on requests made by the MBE:

\begin{hitemize}
	\item Historical data on reservations: $\texttt{mge1}:\vec{p} = (0.18,\ 0.7,\ 0.06,\ 0.06)$,
	\item Pairs only: $\texttt{mge2}:\vec{p} = (0,\ 1,\ 0,\ 0)$,
	\item Singletons and pairs: $\texttt{mge3}:\vec{p} = (0.2,\ 0.8,\ 0,\ 0)$,
	\item Pairs and quads: $\texttt{mge4}:\vec{p} = (0,\ 0.5,\ 0,\ 0.5)$.
\end{hitemize}

We solve the integer programming models in (\ref{model1}) and in (\ref{model2}) both for the Grand Room and the Small Room, where we considered two scenarios for the set of consecutive shows: $V = \{1\}$ (single show) and $V = \{1,2\}$ (double show). Both the basic versions of both models (\texttt{vanilla}), and the versions with the speedup techniques (\texttt{speedup}) are considered and compared. The \texttt{speedup} version is implemented by adding all inequalities described in Section~\ref{sec:speedup} to the ILP formulations. We set $\varepsilon = 0.02$ for the constraints in (\ref{cond:profile}), as we empirically observe this choice for $\varepsilon$ to give a suitable tradeoff between solver performance and solution structure with respect to target profiles. The same forbidden areas are considered for both theatre rooms, as the interseat distances coincide for both rooms. Additionally, we adapt the Gurobi parameters \texttt{Symmetry}, \texttt{Cuts} and \texttt{Presolve} to their aggressive settings. Table~\ref{table:profile_gz} and Table~\ref{table:profile_kz} provide the densities $d(\mathcal{A}(\vec{p}))$ for the Grand Room and the Small Room respectively, where $\mathcal{A}(\vec{p})$ is an optimal safe seating arrangement with respect to an indicated target profile $\vec{p}$, both for the single show and double show case.
\begin{table}[H]
	\TABLE
	{Densities $d(\mathcal{A}(\vec{p}))$ (in \%) of maximum safe seating arrangements $\mathcal{A}(\vec{p})$ in the Grand Room, according to the target profiles. The reported numbers in the columns \texttt{vanilla} and \texttt{speedup} represent time in seconds (rounded to two decimal places).\label{table:profile_gz}}
	{\begin{tabularx}{\textwidth}{@{}cCCCCCC@{}}
		\toprule
		\multicolumn{1}{c}{Target profile} & \multicolumn{3}{c}{Single show} & \multicolumn{3}{c}{Double show}\\
		\cmidrule(r{4pt}){1-7}
		& \multicolumn{1}{c}{Density} & \multicolumn{1}{c}{\texttt{vanilla}} & \multicolumn{1}{c}{\texttt{speedup}} & \multicolumn{1}{c}{Density} & \multicolumn{1}{c}{\texttt{vanilla}} & \multicolumn{1}{c}{\texttt{speedup}}\\
		\cmidrule(l){2-4} \cmidrule(l){5-7}
		$\texttt{mge1}$ & 32 & 3.39 & 1.50 & 63 & 532.69 & 48.28\\
		$\texttt{mge2}$ & 29 & 0.28 & 0.10 & 56 & 6.67 & 2.49\\
		$\texttt{mge3}$ & 30 & 1.39 & 0.97 & 58 & 2107.68 & 6.05\\
		$\texttt{mge4}$ & 36 & 5.29 & 1.10 & 70 & 4485.33 & 726.11\\
		\midrule
		\bottomrule
	\end{tabularx}}
	{}
\end{table}

\begin{table}[H]
	\TABLE
	{Densities $d(\mathcal{A}(\vec{p}))$ (in \%) of maximum safe seating arrangements $\mathcal{A}(\vec{p})$ in the Small Room, according to the target profiles. The reported numbers in the columns \texttt{vanilla} and \texttt{speedup} represent time in seconds (rounded to two decimal places).\label{table:profile_kz}}
	{\begin{tabularx}{\textwidth}{@{}cCCCCCC@{}}
		\toprule
		\multicolumn{1}{c}{Target profile} & \multicolumn{3}{c}{Single show} & \multicolumn{3}{c}{Double show}\\
		\cmidrule(r{4pt}){1-7}
		& \multicolumn{1}{c}{Density} & \multicolumn{1}{c}{\texttt{vanilla}} & \multicolumn{1}{c}{\texttt{speedup}} & \multicolumn{1}{c}{Density} & \multicolumn{1}{c}{\texttt{vanilla}} & \multicolumn{1}{c}{\texttt{speedup}}\\
		\cmidrule(l){2-4} \cmidrule(l){5-7}
		$\texttt{mge1}$ & 34 & 1.19 & 0.41 & 64 & 8.18 & 13.82\\
		$\texttt{mge2}$ & 31 & 0.02 & 0.02 & 58 & 0.30 & 0.35\\
		$\texttt{mge3}$ & 31 & 0.22 & 0.07 & 59 & 2.19 & 0.89\\
		$\texttt{mge4}$ & 37 & 0.08 & 0.11 & 70 & 5.46 & 9.17\\
		\midrule
		\bottomrule
	\end{tabularx}}
	{}
\end{table}

Let us first comment on the densities found in Tables~\ref{table:profile_gz} (Grand Room) and \ref{table:profile_kz} (Small Room). For each of the four target profiles, the differences in density between the Grand Room and the Small Room are small, both for the single show and for the double show situation. This is to be expected as the interseat distances from Figure~\ref{fig:afmetingen} apply to both rooms. Also, in case of a single show, the densities found are rather similar for the four different target profiles, with the exception of \texttt{mge4}. Target profile \texttt{mge4} has a relatively large fraction of families of size 4 (the largest family size considered), which is beneficial for finding seating arrangements with a large density (see the discussion in Section~\ref{sec:building}). However, in case of a single show, all profiles allow a density of around 33\% - this corresponds to a setting with 1/3 of the seats being occupied in a single show.

When analyzing the outcomes for a double show, we observe that the presence of large families (\texttt{mge4}) leads to better densities  - this effect is more pronounced compared to a single show. Another interesting observation is that the densities almost double when compared to a single show. Hence, the effect of the constraint that a seat can be used at most once in two shows is negligible; in other words, the model is able to find two single show seating arrangements with no seats in common such that the numbers of seats occupied in both shows is (almost) balanced. For the target profile based on historical data, \texttt{mge1}, the model is able to find seating arrangements that use almost 2/3 of the available seats. This is an important finding as it gives the MBE an idea of the consequences of having consecutive shows.

Let us now comment on the computation times. In particular, we see that adding a second show to the model drastically increases the computation time of the solver, which can probably be explained by the fact that additional symmetries are introduced in the problem by adding a second show and that the number of variables and constraints both increase linearly in $|\mathcal{S}|$. Furthermore, the choice for the target profile also largely influences the running time of the algorithm. It is striking to see that the instances for which the algorithm has the worst performance are also the ones for which the target profiles are further away from intuitively optimal, i.e. relatively large proportions of small families and relatively small proportions of large families. For the instances on the Grand Room, we see that the impact of adding the speedup techniques is rather large. This can be explained by the fact that these instances have a rich variety of symmetries.

\subsection*{Real life implementation: alternating empty rows}
Recall that in Section~\ref{sec:hilbert}, we analyzed a setting where in each show, seats in row $r$ could be occupied by guests whenever row $r-1$ and $r+1$ (if existent) were empty, for any row $r$. 

Tables~\ref{table:omenom_profile_gz} and \ref{table:omenom_profile_kz} report the optimum densities of safe seating arrangements $\mathcal{A}'(\vec{p})$ that have the property of using only one of two consecutive rows at a time per show, for the Grand Room and the Small Room respectively. For the double show case, this means in the first show, all odd-numbered rows are used, and all even-numbered rows in the second. We only report computation times for the \texttt{vanilla} descriptions, as these speedup techniques now only yield redundant inequalities. The column called ``Loss (\%)'' indicates the percentual loss of occupied seats, which can be seen as a proxy for the loss in revenue.

\begin{table}[H]
	\TABLE
	{Densities $d(\mathcal{A}'(\vec{p}))$ (in \%) of maximum safe seating arrangements $\mathcal{A}'(\vec{p})$ in the Grand Room, according to the four target profiles when one of every two consecutive rows might be occupied by guests.\label{table:omenom_profile_gz}}
	{\begin{tabularx}{\textwidth}{@{}cCCCCCC@{}}
		\toprule
		\multicolumn{1}{c}{Target profile} & \multicolumn{3}{c}{Single show} & \multicolumn{3}{c}{Double show}\\
		\cmidrule(r{4pt}){1-7}
		& \multicolumn{1}{c}{Density} &
		\multicolumn{1}{c}{Loss (\%)} &
		\multicolumn{1}{c}{\texttt{vanilla} (s)} &
		
		\multicolumn{1}{c}{Density} &
		\multicolumn{1}{c}{Loss (\%)} &
		\multicolumn{1}{c}{\texttt{vanilla} (s)}\\
		\cmidrule(l){2-4} \cmidrule(l){5-7}
		$\texttt{mge1}$ & 29 & -8.5 & 0.15 & 57 & -9.4 & 0.33\\
		$\texttt{mge2}$ & 27 & -7.8 & 0.01 & 52 & -7.5 & 0.05\\
		$\texttt{mge3}$ & 27 & - 9.1 & 0.05 & 52 & -9.8 & 0.07 \\
		$\texttt{mge4}$ & 34 & -5.8 & 0.04 & 65 & -6.0 & 0.12 \\
		\midrule
		\bottomrule
	\end{tabularx}}
    {}
\end{table}

\begin{table}[H]
	\TABLE
	{Densities $d(\mathcal{A}'(\vec{p}))$ (in \%) of maximum safe seating arrangements $\mathcal{A}'(\vec{p})$ in the Small Room, according to the four target profiles when one of every two consecutive rows might be occupied by guests.\label{table:omenom_profile_kz}}
	{\begin{tabularx}{\textwidth}{@{}cCCCCCC@{}}
		\toprule
		\multicolumn{1}{c}{Target profile} & \multicolumn{3}{c}{Single show} & \multicolumn{3}{c}{Double show}\\
		\cmidrule(r{4pt}){1-7}
		& \multicolumn{1}{c}{Density} &
		\multicolumn{1}{c}{Loss (\%)} &
		\multicolumn{1}{c}{\texttt{vanilla} (s)} &
		
		\multicolumn{1}{c}{Density} &
		\multicolumn{1}{c}{Loss (\%)} &
  	    \multicolumn{1}{c}{\texttt{vanilla} (s)}\\
		\cmidrule(l){2-4} \cmidrule(l){5-7}
		$\texttt{mge1}$ & 30 & -15.7 & 0.06 & 58 & -9.8 & 0.11 \\
		$\texttt{mge2}$ & 26 & -14.8 & 0.00 & 52 & -9.6 & 0.01 \\
		$\texttt{mge3}$ & 26 & -16.8 & 0.02 & 52 & -11.8 & 0.08 \\
		$\texttt{mge4}$ & 34 & -13.5 & 0.01 & 66 & -5.7 & 0.03\\
		\midrule
		\bottomrule
	\end{tabularx}}
	{}
\end{table}

Clearly, as the results in Tables~\ref{table:omenom_profile_gz} and \ref{table:omenom_profile_kz} correspond to a more restricted setting of our problem, the realized densities are always smaller than those achieved for the setting where all rows can be used for all shows. Indeed, we observe that for all instances, especially the ones based on the Small Room, the percentual loss of occupied seats is rather significant, with no percentual loss smaller than 5.7\% of occupied seats. 

Computation times for this setting are much smaller. This is caused by the much smaller size of the resulting instances and much fewer dependencies between the variables. 

The solutions that correspond to the occupancies for the target profile \texttt{mge1} given in Tables~\ref{table:profile_gz} to~\ref{table:omenom_profile_kz} are provided in the appendices. We used the color red to indicate that those seats are forbidden for use by guests and the color green (and blue for two consecutive shows) to indicate seats that can be occupied by guests. 


\section{Upper bounds for the occupancy in MBE}
\label{sec:ub}

In Table~\ref{table:virtual_seats}, we list the number of seats $|\mathcal{S}|$ and the number of virtual seats (the ``rim'') $|(\mathcal{S}+\mathcal{T})\setminus \mathcal{S}|$, in each of the two rooms of the MBE. By applying Theorems~\ref{thm:volume_bound} and~\ref{thm:bound2} to the rooms of the MBE, we can find upper bounds on the achievable occupancy. 

\begin{table}[htbp]
    
	\TABLE
	{Number of (virtual) seats in the Grand Room and the Small Room.\label{table:virtual_seats}}
	{\begin{tabularx}{0.7\textwidth}{@{}CCC@{}}
		\toprule
		& Grand Room & Small Room\\
		\midrule
		$|\mathcal{S}|$ & 1250 & 400\\
		$|(\mathcal{S}+\mathcal{T})\setminus \mathcal{S}|$ & 458 & 133\\
		\bottomrule
	\end{tabularx}}
	{}
\end{table}

From Theorem~\ref{thm:volume_bound}, we can deduce the following bound for a single safe seating arrangement $\mathcal{A}$ for the Small Room:
$$\sum_{t \in T} (2t+3) n_t(\mathcal{A}) = 5n_1(\mathcal{A}) + 7n_2(\mathcal{A})+9n_3(\mathcal{A})+11n_4(\mathcal{A}) \le 400 + 135 = 533.$$
 Analogously, we have for the Grand Room
$$5n_1(\mathcal{A}) + 7n_2(\mathcal{A})+9n_3(\mathcal{A})+11n_4(\mathcal{A}) \le 1250 + 458 = 1708.$$

We consider these volume bounds on the realized safe seating arrangements $\mathcal{A}_{max}$ corresponding to the data in Table~\ref{table:profile_gz} and Table~\ref{table:profile_kz} on the single show setting, for the Grand Room and the Small Room respectively. The following table provides the left hand sides on these volume bounds to illustrate the strength of these bounds. Recall the right hand sides for the Grand Room (1708) and Small Room (533). 
\begin{table}[htbp]
	\TABLE
	{Left hand sides of the volume bounds of Theorem~\ref{thm:volume_bound} for the different target profiles on the MBE theatre rooms.\label{table:bounds1}}
	{\begin{tabularx}{0.65\textwidth}{@{}ccc@{}}
		\toprule
		\multicolumn{1}{c}{Target profile} & \multicolumn{2}{c}{Theorem~\ref{thm:volume_bound} LHS}\\
		\cmidrule(r{4pt}){1-1} \cmidrule(l){2-3}
		& Grand Room & Small Room\\
		\midrule
		\texttt{mge1} & 1389 & 466\\
		\texttt{mge2} & 1260 & 427\\
		\texttt{mge3} & 1359 & 457\\
		\texttt{mge4} & 1340 & 443\\
		\midrule
		\bottomrule
	\end{tabularx}}
	{}
\end{table}
The interpretation of the numbers depicted in Table~\ref{table:bounds1} is the number of seats in $\mathcal{S} + \mathcal{T}$ that are covered by the corresponding trapezoid packing. We observe that approximately 75 to 85 \% of (virtual) seats are covered by a trapezoid. The gap on the volume bound can be explained by the fact that the trapezoid forms do not allow for a perfect packing of seats, and this effect is amplified by the inclusion of target profiles, which further restrict the set of possible trapezoid packings.

\ignore{
It turns out that the bounds from Theorem~\ref{thm:volume_bound} are rather loose bounds on the seating arrangements $\mathcal{A}_{max}$ corresponding to the model (\ref{model1}). This can partly be explained due to the fact that both the Grand Room and the Small Room consist of small components such as balconies with fewer than five rows. Therefore the fraction of virtual seats $\frac{|(\mathcal{S}+\mathcal{T})\setminus \mathcal{S}|}{|\mathcal{S}|}$ is rather large compared to our example of the Hilbert theatre.}

Secondly, we consider the bound given by Theorem~\ref{thm:bound2}. Now, we consider the upper bound for occupation densities $d(\mathcal{A}(\vec{p}))$ of this bound corresponding to maximum safe seating arrangements in the model of (\ref{model2}). For convenience, we include the results on the realized occupation densities for the single show setting using all rows again in Table~\ref{table:bounds2}.

\begin{table}[htbp]
	\TABLE
	{Upper bounds on the occupation density $d(\mathcal{A}(\vec{p}))$ (in \%), where $\mathcal{A}(\vec{p})$ is a maximum density safe seating arrangement satisfying target profile $\vec{p}$, together with actually realized densities.\label{table:bounds2}}
	{\begin{tabularx}{0.87\textwidth}{@{}cccc@{}}
		\toprule
		\multicolumn{1}{c}{Target profile} & \multicolumn{3}{c}{Occupation density} \\
		\cmidrule(r{4pt}){1-1} \cmidrule(l){2-4}
		 & UB $d(\mathcal{A}(\vec{p}))$ & $d(\mathcal{A}(\vec{p}))$ (Grand Room) & $d(\mathcal{A}(\vec{p}))$ (Small Room)\\
		\midrule
		\texttt{mge1} & 37.5 & 32 & 34\\
		\texttt{mge2} & 38.8 & 29 & 31\\
		\texttt{mge3} & 36.4 & 30 & 31\\
		\texttt{mge4} & 43.5 & 36 & 37\\
		\midrule
		\bottomrule
	\end{tabularx}}
	{}
\end{table}

We observe that the realized optimal densities are not very far away from the respective upper bounds in Table~\ref{table:bounds2}. Nevertheless, the bound of Theorem~\ref{thm:bound2} is based on perfect tilings of trapezoids of one family size in a suitable chosen theatre architecture, which is not the case for typical inputs. This is a possible explanation for the gap between the bound and the realized density


\section{Conclusion}
\label{sec:conclusion}

The 1.5 meter constraint has a huge impact on the occupancy when filling a theatre. In case of the MBE, when performing a single show on an evening, occupancy will not exceed 40\% (both for the Grand Room and the Small Room). However, allowing two shows per evening, it is possible to reach an occupancy of 70\% while satisfying the constraint that no seat is sold twice. A more logistically suitable solution is to use alternating empty rows, but this comes at the cost of losing at least 5\% on the number of occupied seats. The corresponding solutions, together with other innovations, may offer some hope to theatres to remain competitive.\\[2ex]

\noindent
{\bfseries Acknowledgements:} {\em Omitted for anonymization}\ignore{We thank Mici Fiedeldij, Tomas Tielemans, and Marcia van den Wildenberg (MBE) for the pleasant cooperation, and we thank Laetitia Ouillet for organizing the connection.}

\newpage
\begin{APPENDICES}
\section{Solutions for \texttt{mge1} (all rows)}
The labels in the figures indicate the segments within the theatre rooms that are depicted.
\begin{figure*}[htb]
    \centering
    \subfloat[Maximum number of occupied seats (in green) of the Small Room for a single show with target profile \texttt{mge1}.]{%
      \includegraphics[scale = 0.15]{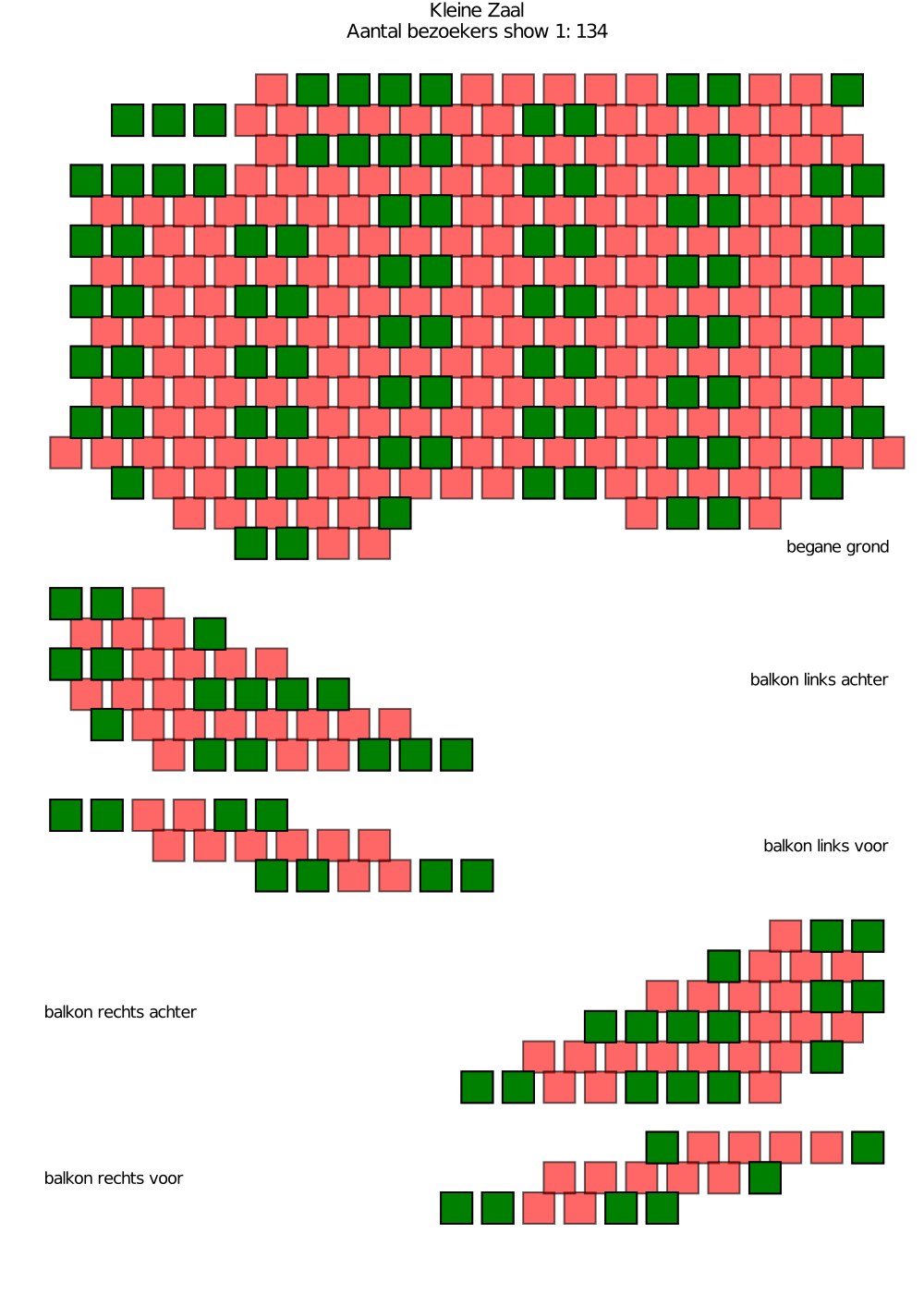}%
    }%
    \qquad
    \subfloat[Maximum number of occupied seats (in green and blue) of the Small Room for two consecutive shows with target profile \texttt{mge1}.]{%
      \includegraphics[scale = 0.15]{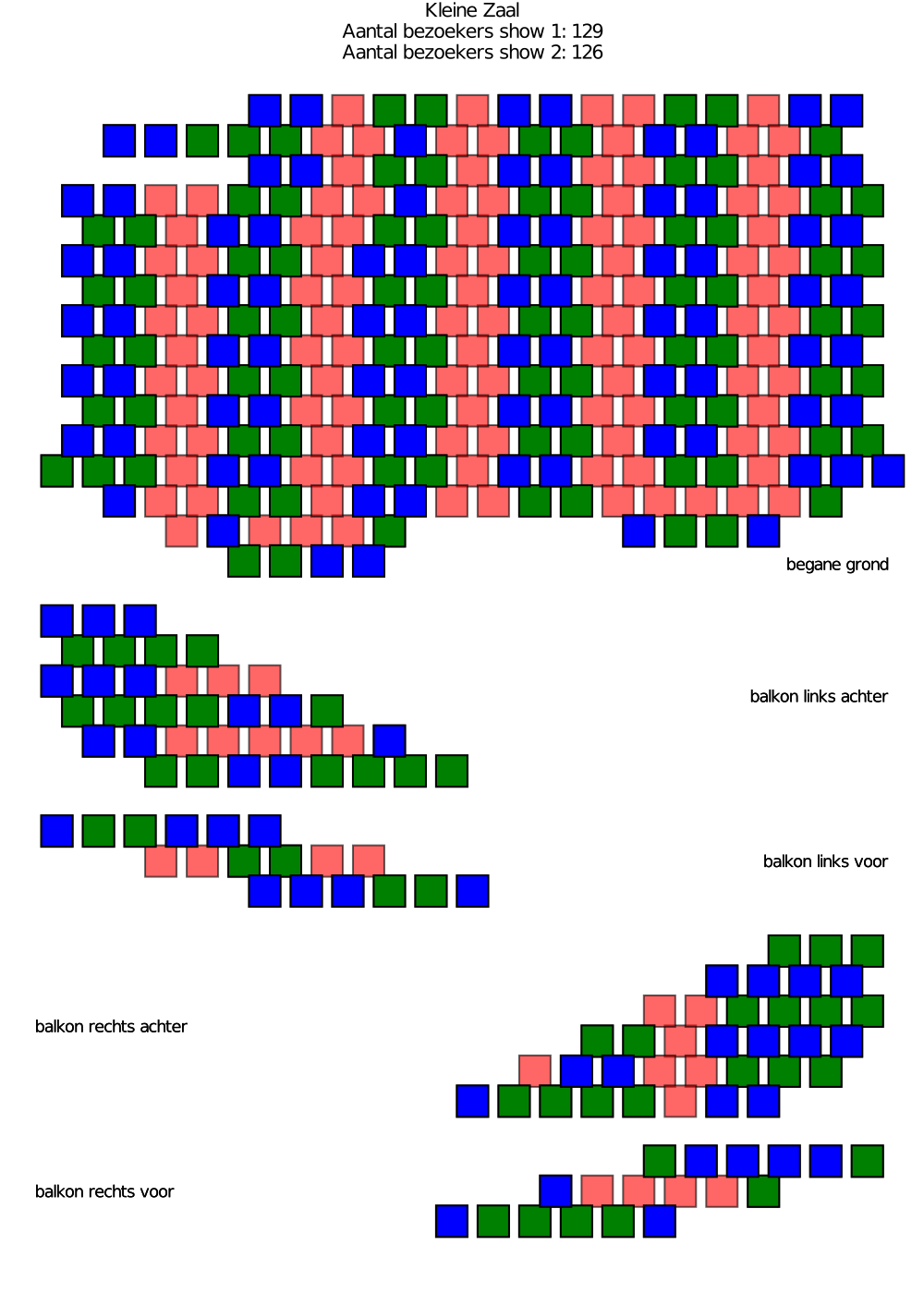}%
    }

    \subfloat[Maximum number of occupied seats (in green) of the Grand Room for a single show with target profile \texttt{mge1}.]{%
      \includegraphics[scale = 0.09]{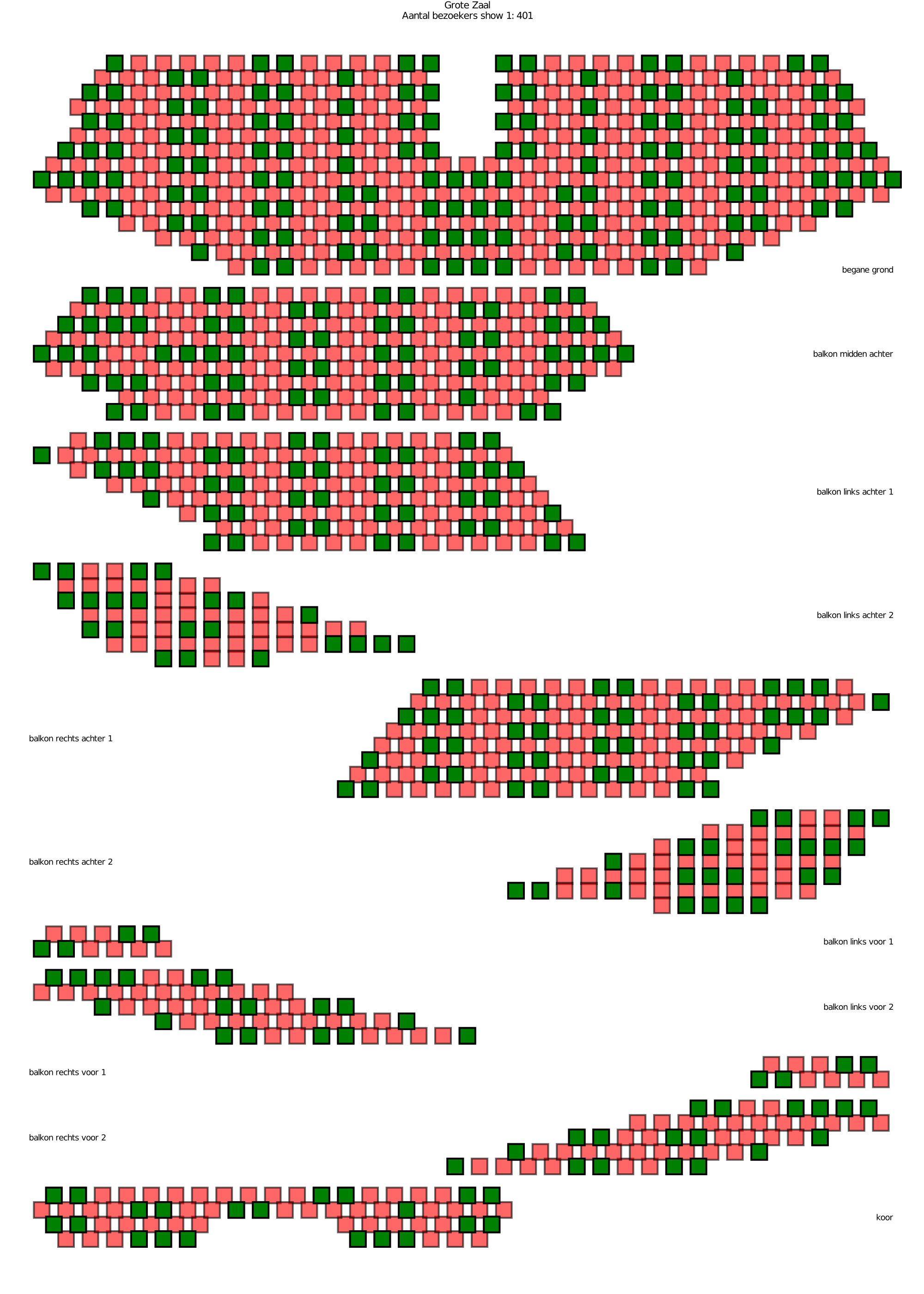}%
    }%
    \qquad
    \subfloat[Maximum number of occupied seats (in green and blue) of the Grand Room for two consecutive shows with  target profile \texttt{mge1}.]{%
      \includegraphics[scale = 0.09]{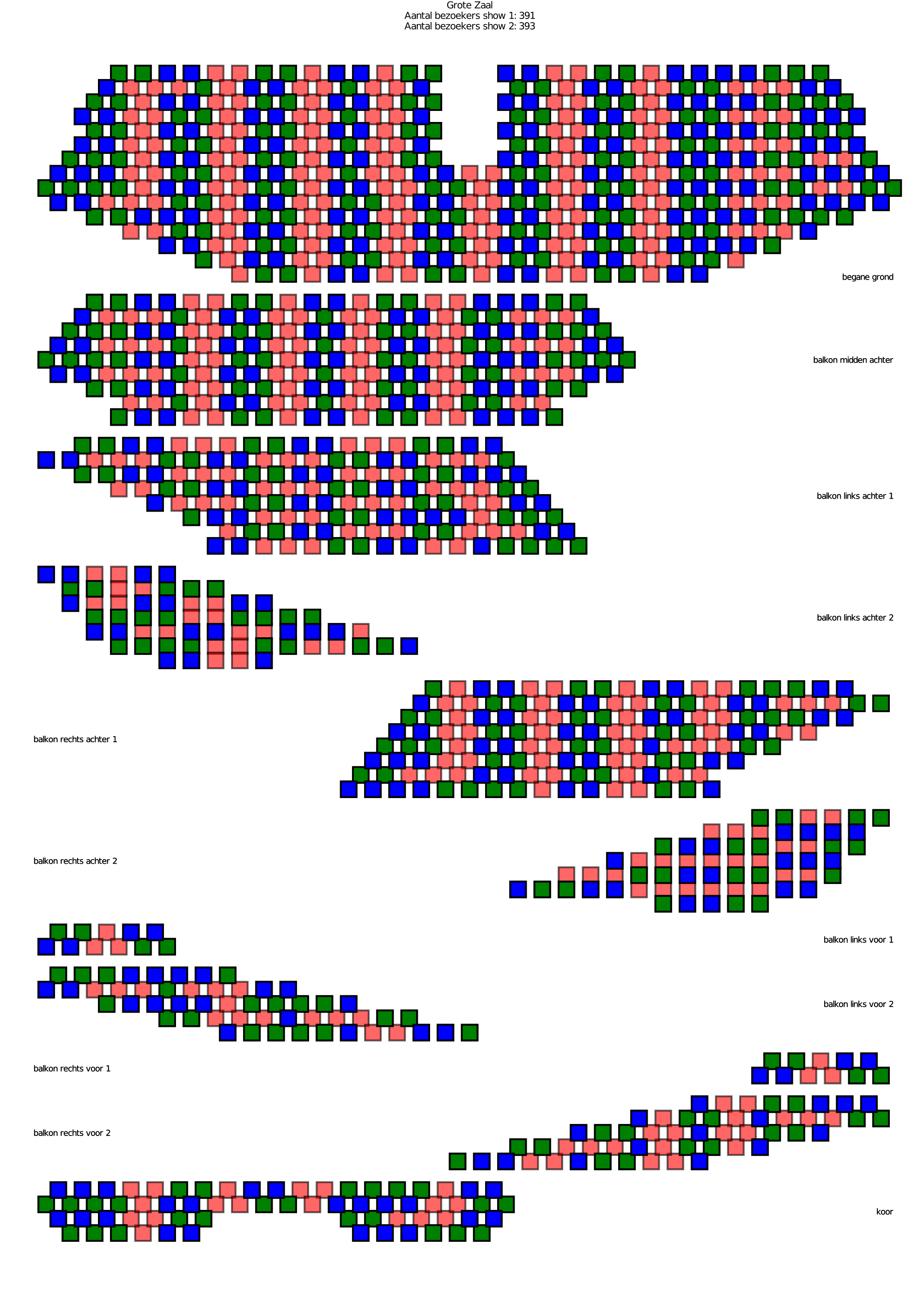}%
    }
    \label{fig:floor_plans_mge1}
\end{figure*}

\section{Solutions for \texttt{mge1} (occupied rows are between two empty rows)}
The labels in the figures indicate the segments within the theatre rooms that are depicted.
\begin{figure*}[htb]
    \centering
    \subfloat[Maximum number of occupied seats (in green) of the Small Room for a single show with target profile \texttt{mge1}.]{%
      \includegraphics[scale = 0.14]{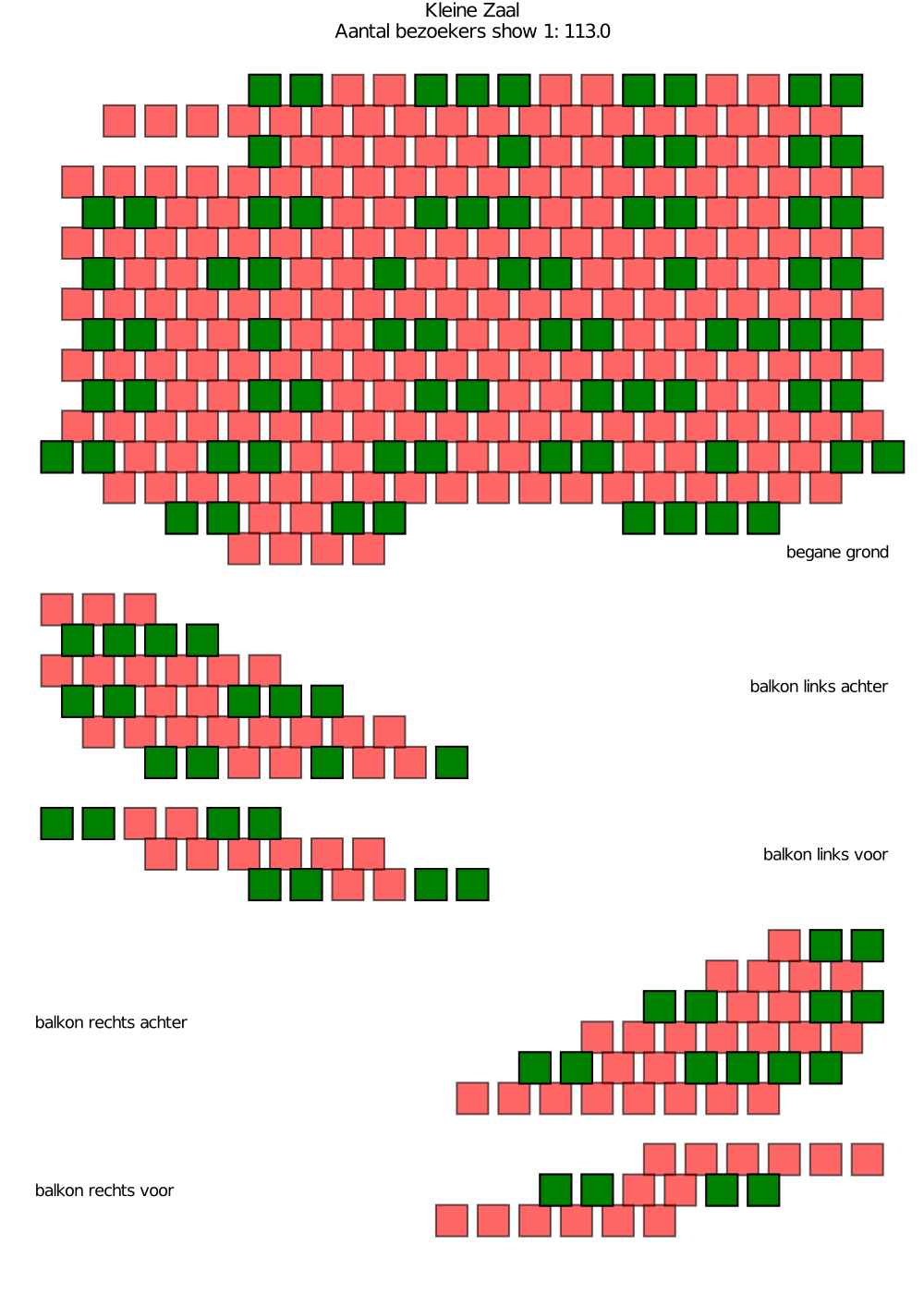}%
    }%
    \qquad
    \subfloat[Maximum number of occupied seats (in green and blue) of the Small Room for two consecutive shows with target profile \texttt{mge1}.]{%
      \includegraphics[scale = 0.14]{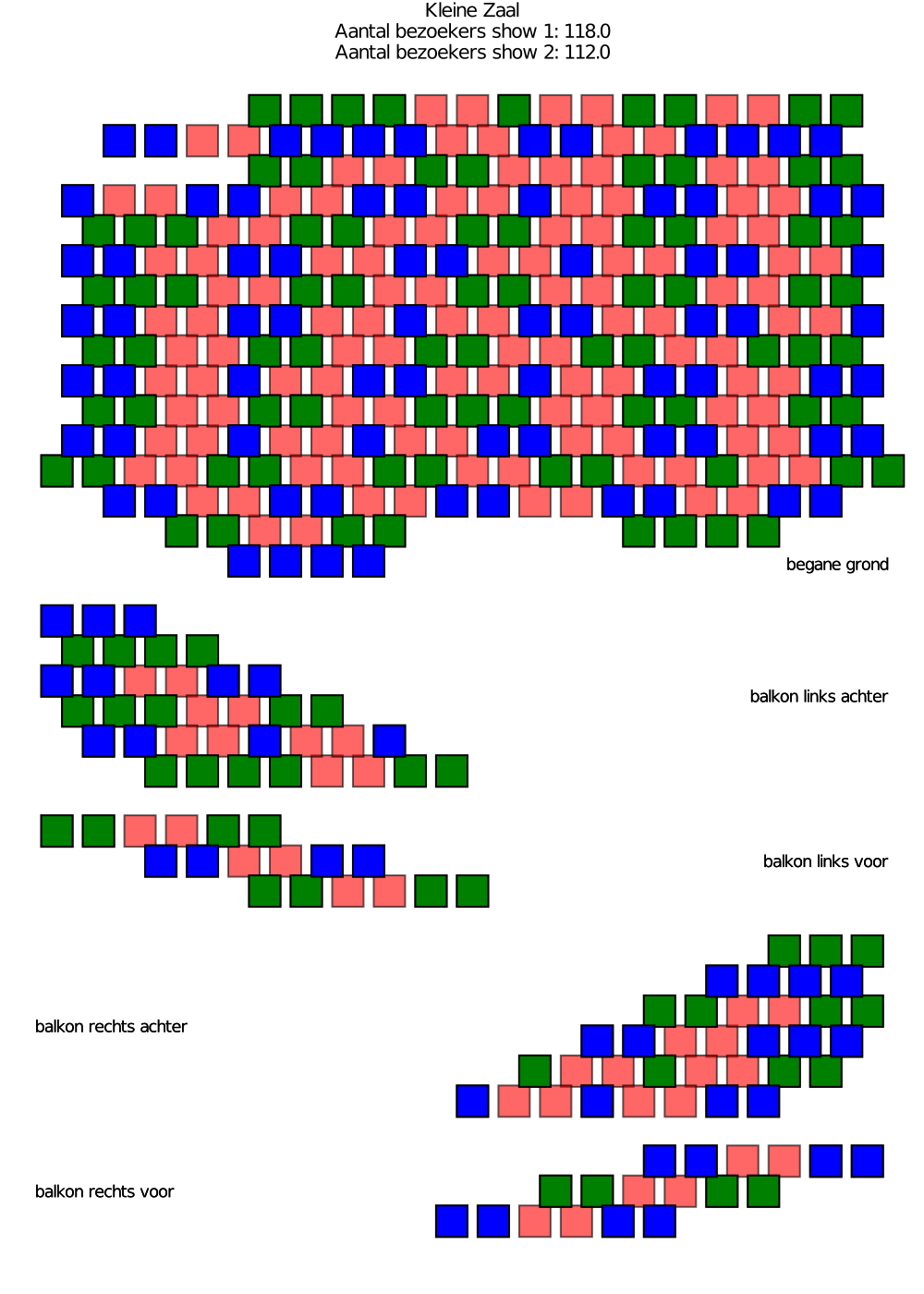}%
    }%

    \subfloat[Maximum number of occupied seats (in green) of the Grand Room using a single show with the target profile \texttt{mge1}.]{%
      \includegraphics[scale = 0.09]{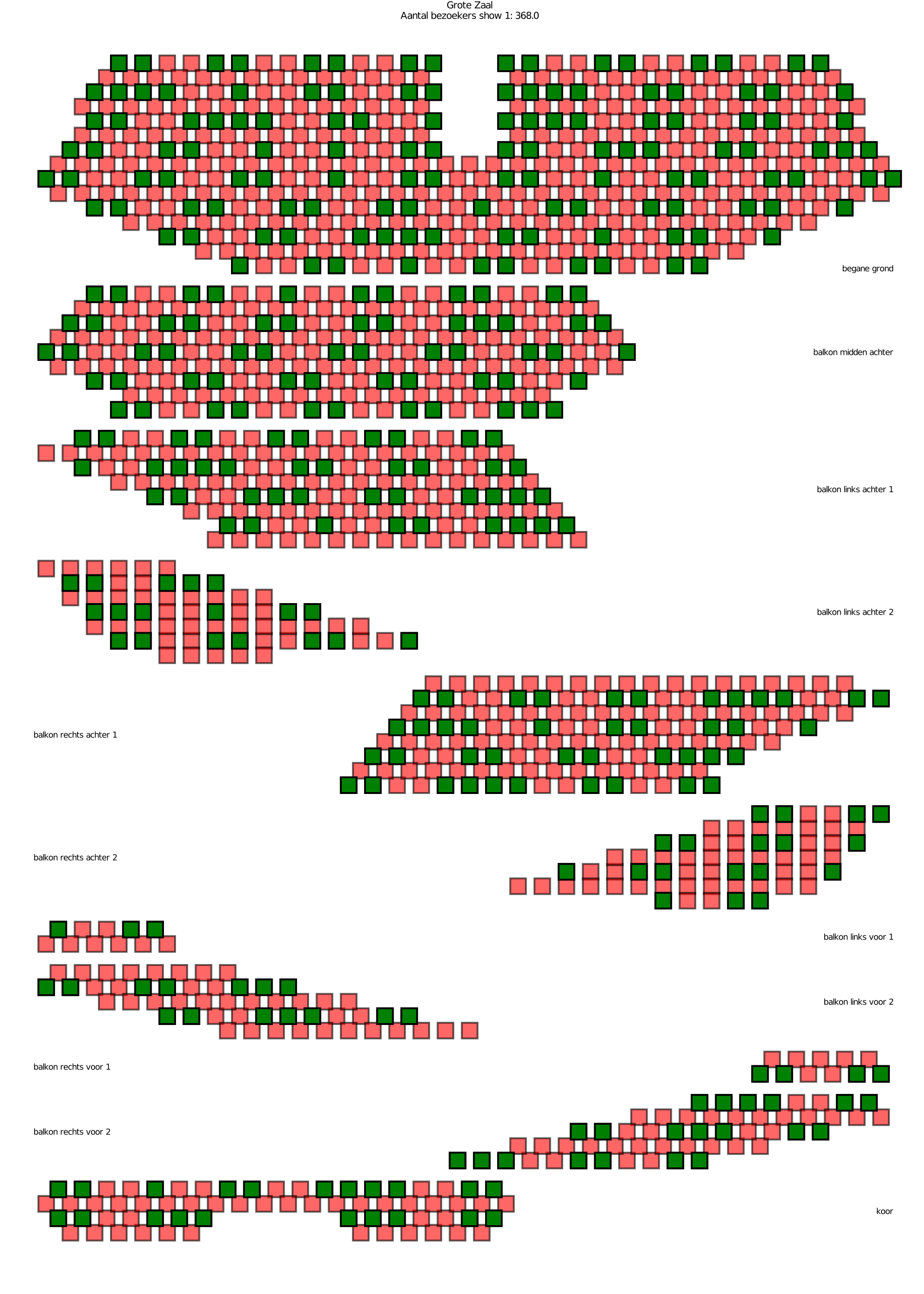}%
    }%
    \qquad
    \subfloat[Maximum number of occupied seats (in green and blue) of the Grand Room using two consecutive shows with the target profile \texttt{mge1}.]{%
      \includegraphics[scale = 0.09]{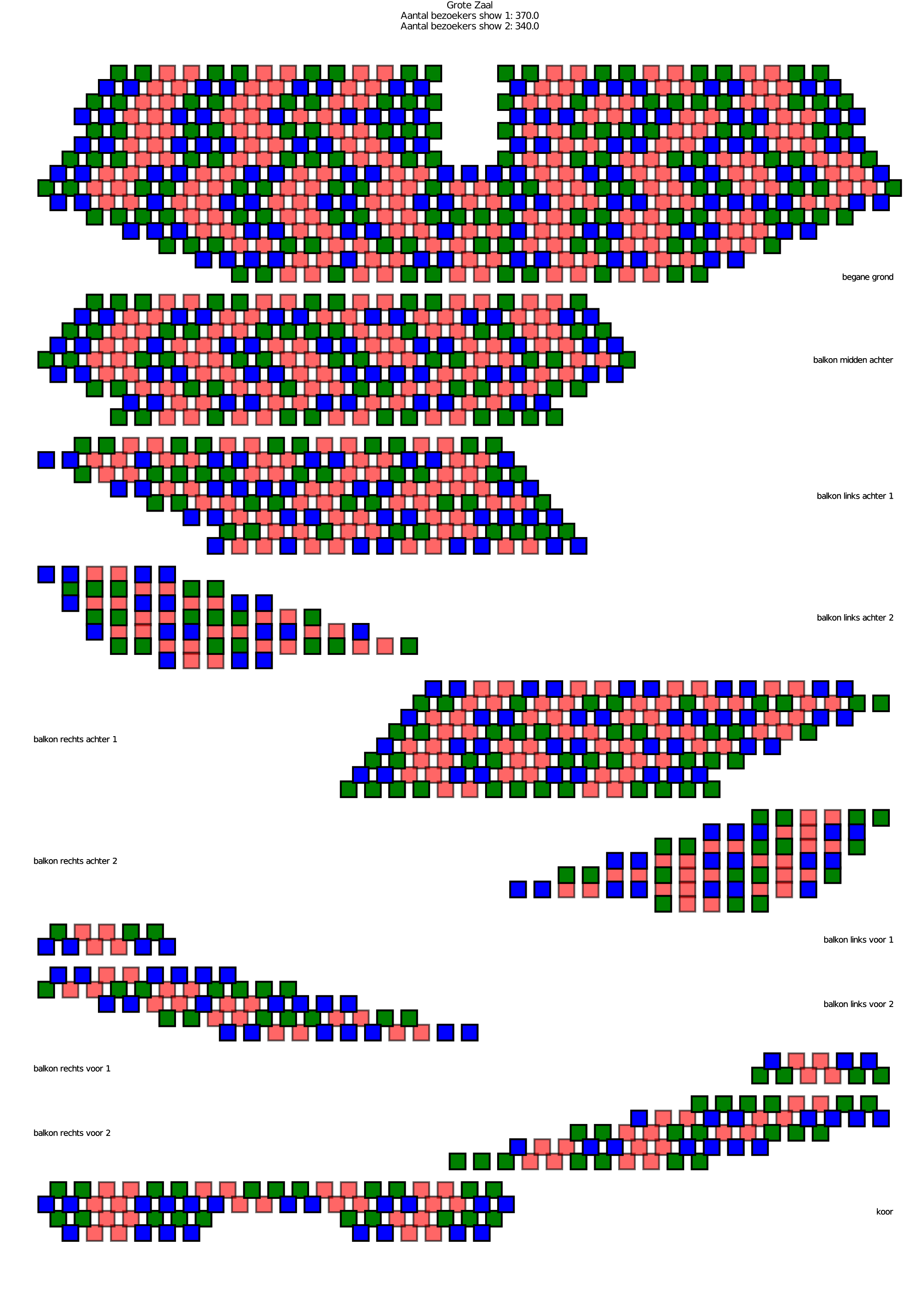}%
    }%
    \label{fig:floor_plans_mge1_alternating}
\end{figure*}
\end{APPENDICES}
\end{document}